\documentclass[pdflatex,sn-mathphys-num]{sn-jnl}%

\usepackage{graphicx}%
\usepackage{multirow}%
\usepackage{amsmath,amssymb,amsfonts}%
\usepackage{amsthm}%
\usepackage{mathrsfs}%
\usepackage[title]{appendix}%
\usepackage{xcolor}%
\usepackage{textcomp}%
\usepackage{manyfoot}%
\usepackage{booktabs}%
\usepackage{multirow}
\usepackage{algorithmicx}%
\usepackage{algpseudocode}%
\usepackage{listings}%
\usepackage{algorithm2e}

\usepackage{tikz}
\usetikzlibrary{positioning, backgrounds, fit}
\usepackage{subcaption}
\usepackage{url}
\usepackage{hyperref}
\usepackage[capitalize]{cleveref}

\usetikzlibrary{shapes.geometric, arrows}
\usepackage{booktabs}
\usepackage{multirow}

\usepackage[dvipsnames]{xcolor}

\usepackage{colortbl}
\colorlet{colorFst}{Green!15}       %
\newcommand{\fs}{\rowcolor{colorFst}}   %

\tikzstyle{startstop} = [rectangle, rounded corners, minimum width=3cm, minimum height=1cm, text centered, draw black, fill=red!30]

\definecolor{safetyColor}{HTML}{0d6a82}
\definecolor{biodiversitycolor}{HTML}{fb4d3d}
\definecolor{3dmodelColor}{HTML}{345995}
\definecolor{applicationColor}{HTML}{e7901a}
\definecolor{backgroundColor}{HTML}{2a9d8f}
\definecolor{challengeColor}{HTML}{606c38}
\definecolor{applicationBackground}{HTML}{ffb703}

\tikzset{
 base node/.style={
 minimum width=1cm,
 align=center,
 minimum height=0.75cm,
 inner sep=5pt
 },
 category box/.style={
 rectangle,
 rounded corners,
 draw=black,
 inner sep=5pt
 },
 arrow/.style={
 -latex,
 ultra thick,
 opacity=0.9
 },
 3dmodels/.style={base node, fill=white, draw=black},
 biodiversity/.style={base node, fill=biodiversitycolor!5, draw=biodiversitycolor},
 safety/.style={base node, fill=safetyColor!5, draw=safetyColor},
 applications/.style={base node, fill=applicationColor!5, draw=applicationColor},
 title/.style={minimum height=0cm, align=center},
 optimized/.style={dashed, draw=safetyColor, thick},
 optimizedbox/.style={optimized, fill=safetyColor!5}
}

\theoremstyle{thmstyleone}%

\theoremstyle{thmstyletwo}%

\theoremstyle{thmstylethree}%

\begin{document}

\title[Physics-Informed Neural Networks for Thermophysical Property Retrieval]{Physics-Informed Neural Networks for Thermophysical Property Retrieval}

\author[1]{\fnm{Ali} \sur{Waseem}}\email{ali.waseem607@gmail.com}
\equalcont{These authors contributed equally to this work.}
\author*[1]{\fnm{Malcolm} \sur{Mielle}}\email{pinnit@mielle.dev}
\equalcont{These authors contributed equally to this work.}
\affil[1]{\orgdiv{Schindler EPFL Lab}, \orgname{Schindler}, \orgaddress{\street{Quartier de l'innovation}, \city{Lausanne}, \postcode{1015}, \state{Vaud}, \country{Switzerland}}}

\abstract{

    Inverse heat problems refer to the estimation of material thermophysical properties given observed or known heat diffusion behaviour.
Inverse heat problems have wide-ranging uses, but a critical application lies in quantifying how building facade renovation reduces thermal transmittance, a key determinant of building energy efficiency.
However, solving inverse heat problems with non-invasive data collected in situ is error-prone due to environmental variability or deviations from theoretically assumed conditions.
Hence, current methods for measuring thermal conductivity are either invasive, require lengthy observation periods, or are sensitive to environmental and experimental conditions.
Here, we present a PINN-based iterative framework to estimate the thermal conductivity $k$ of a building wall from a set of thermographs; our framework alternates between estimating the forward heat problem with a PINN for a fixed $k$, and optimizing $k$ by comparing the thermographs and surface temperatures predicted by the PINN, repeating until the estimated $k$'s convergence.
Using both environmental data captured by a weather station and data generated from Finite-Volume-Method analysis software simulations, we accurately predict $k$
across different environmental conditions and data collection sampling times, given the temperature profile of the wall at dawn is close to steady state.
Although violating the steady-state assumption impacts the accuracy of $k$'s estimation, we show that our
proposed framework still only exhibits a maximum MAE of $4.0851$.
Our work demonstrates the potential of PINN-based methods for reliable estimation of material properties in situ and under realistic conditions, without requiring lengthy measurement campaigns.
Given the lack of research on using machine learning, and more specifically on PINNs, for solving in-situ inverse problems, we expect our work to be a starting point for more research on the topic.

}


\maketitle

\section{Introduction}\label{sec:introduction}

Boundary value problems involve partial differential equations (PDEs) constrained by boundary conditions, requiring solutions that satisfy both the governing equations and the imposed constraints.
These forward problems are ubiquitous in science and engineering, in applications such as topology optimization~\cite{wang2004pde, wallin2020Consis}, system design~\cite{ugail1999Techni}, and fluid dynamics~\cite{ferziger2020Comput}.
In contrast, inverse problems focus not on solving the equations themselves but on inferring unknown parameters or functions within the system's governing model.
Inverse problems can be found in fields like geophysics~\cite{ansari2022Solvin}, environmental monitoring~\cite{bagtzoglou2005Near_r}, and medicine~\cite{nolte2022Invers}.
While numerical methods are effective at solving forward and inverse boundary value problems, their computational demands often scale prohibitively with system complexity~\cite{fischer2020Scalab}.
Furthermore, in practice, solving inverse problems is difficult as linear, overdetermined inverse problems are the exception rather than the rule~\cite{cardiff2008Effici}---due, for example, to realistic conditions (such as dynamic environmental conditions) that must be taken into account while estimating the system's parameters.

Heat diffusion---the process governing thermal energy transfer through materials---can be modeled as a boundary-value problem, where the temperature distribution in a material is determined by an initial condition and boundary constraints (e.g., Dirichlet or Neumann conditions).
On the other hand, the inverse heat diffusion problem is concerned with estimating, given observed temperature data, the thermal properties of the material, such as thermal conductivity.
While such inverse methods are widely used in manufacturing~\cite{10178253, 8264402} and quality control~\cite{8310645}, a key application of inverse heat problems lies in estimating the thermal transmittance (U-value) (or its inverse, the thermal insulance R-value, which is equal to thickness of a material divided by its thermal conductivity $k$), of building façades before and after retrofitting, to evaluate the impact of renovations.
Indeed, building façades dictate the majority of energy flux in and out of a building~\cite{Nardi_2018}, and poor insulation exacerbates energy demand, carbon emissions, and occupant discomfort.
Yet, while façade retrofits are capable of reducing total energy demand by 35-40\%~\cite{SARIHI2021102525}, nearly 75\% of the EU's building stock remains energy-inefficient under current standards~\cite{act2011communication}, with only 0.2\% of buildings annually undergoing renovations that reduce energy consumption by at least 60\%~\cite{act2011communication}.
As global temperatures rise, bridging this gap between potential and practice is critical not only for climate mitigation but also for public health and equity.
Usually, the $R$-value of a given material is found through existing tabulated data.
However, tabulated values may introduce errors in the in-situ evaluation of infrastructure~\cite{tejedor_quantitative_2017} since $R$ is influenced not only by density and thickness of the material, but also by the installation quality (gaps or compression in the insulation), aging (material degradation) due to, for example, environmental factors like temperature changes or ultraviolet light.

Current International Organization for Standardization (ISO) methods~\cite{noauthor_iso_nodate, noauthor_iso_nodate-1} for measuring thermal transmittance involve the use of heat-flow meters (HFM), taking several days to estimate thermal transmittance, and can potentially damage the building envelope~\cite{tardy_review_2023}. %
Although HFM is commonly employed, it faces meteorological and operational challenges.
For instance, heat flow plates can interfere with heat flow, thereby affecting measurements~\cite{desogus_comparing_2011, cesaratto_effect_2011, trethowen_measurement_1986}, and the primary cause of inaccuracies in thermal transmittance estimation is improper positioning of heat flow plates~\cite{peng_situ_2008}.
While methods to improve accuracy by post-processing HFM data have been proposed~\cite{evangelisti_methodological_2020}, ISO 9869 still requires lengthy, costly, and intrusive methods to determine R-values in situ~\cite{tardy_review_2023} and does not address the operational complexity of HFM methods.
Indeed, the averaging method for estimating $R$-values takes at least 3 days~\cite{tejedor_quantitative_2017, biddulph_inferring_2014}---and sometimes up to 14 days~\cite{biddulph_inferring_2014}---and audits tend to be limited to early morning measurements with low wind speeds and large indoor/outdoor temperature differences~\cite{tejedor_quantitative_2017, tejedor_assessing_2018, lu_application_2019, Nardi_2018}.
To avoid potential damage and disruptions to building users, non-invasive methods are preferred.

Non-invasive methods mostly use thermography rather than HFM to estimate thermal properties of building facades~\cite{tejedor_assessing_2018, mahmoodzadeh_determining_2021, marino_estimation_2017, lu_application_2019, tejedor_quantitative_2017, GONZALEZAGUILERA201329, VIDERASRODRIGUEZ2024114874, ALBATICI2015218}.
However, current non-invasive methods relying on thermography often require particular environmental conditions for observation and/or long observation periods to provide accurate estimates of thermal transmittance~\cite{tejedor_assessing_2018, mahmoodzadeh_determining_2021, bacher_identifying_2011, biddulph_inferring_2014, marino_estimation_2017, lu_application_2019}, which are not easily applicable in real-life scenarios.
For example, high correlation between variance in thermal transmittance and outer air temperature can limit the accuracy of the estimation and the amount of situations where the method can be used~\cite{tejedor_assessing_2018}.
Furthermore, environmental conditions and thermal imaging artifacts can also impact $R$-value estimation~\cite{mahmoodzadeh_determining_2021}; often, windy and rainy days must be avoided~\cite{marino_estimation_2017} and multiple camera setups must be used to improve accuracy~\cite{marino_estimation_2017}.
To avoid the effect of environmental conditions, measurements can be taken from inside the structure~\cite{tejedor_quantitative_2017} and a quasi-steady state must often be assumed~\cite{tejedor_quantitative_2017}.
Thus, while the thermal resistance of a material is an important parameter for heat flux simulation, it is difficult to obtain in a reliable and accurate manner in-situ---e.g. recent research aiming to bridge the gap between computer vision and Finite-Element Analysis (FEA)~\cite{chassaing2025thermoxels} to evaluate the quality of building facade still relies on a known estimation of material properties.

To improve the adaptability and scalability of determining the solution of PDEs and boundary problems, recent advances in machine learning have proposed using neural networks to estimate PDE solutions by leveraging knowledge of the physical system being estimated---such models are referred to as physics-informed neural networks (PINNs)~\cite{raissi_physics-informed_2019}.
PINNs have gained recognition as a novel framework for integrating physics-based domain knowledge into deep learning models, especially with regard to addressing complex forward and inverse problems~\cite{bdcc6040140}.
PINNs have been applied to solve problems in diverse application domains such as weather modeling~\cite{hu2023predicting}, epidemiology~\cite{millevoi2024physics}, fluid dynamics~\cite{chiu2022can}, and daylight simulation in buildings~\cite{labib_utilizing_2025}.
PINN-based methods have been used to solve heat transfer problems under more realistic conditions where traditional methods fall short~\cite{cai_physics-informed_2021}---for example, PINNs have been used to address conductive heat transfer problems in manufacturing and engineering, estimating transient temperature profiles successfully~\cite{zobeiry_physics-informed_2021, billah_physics-informed_2023}.
However, state-of-the-art PINN methods for heat diffusion need known material properties~\cite{zobeiry_physics-informed_2021}, and most rely on data collected in controlled environments~\cite{billah_physics-informed_2023}, limiting their applicability.
Thus, practical challenges persist, such as the need for known material properties during training~\cite{zobeiry_physics-informed_2021} or the difficulty of obtaining precise input data in realistic scenarios~\cite{billah_physics-informed_2023}.

\begin{figure}[t]
    \begin{subfigure}{\textwidth}
        \centering
        \scalebox{0.8}{\begin{tikzpicture}
                \input{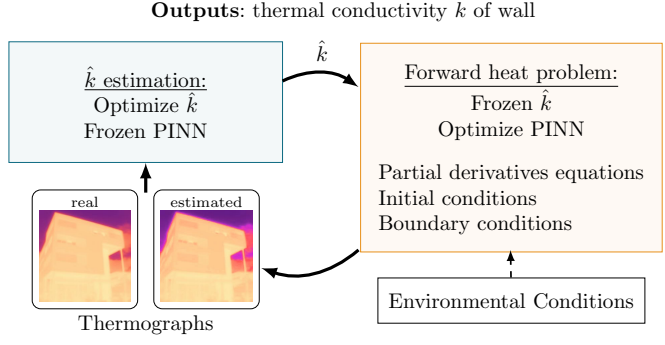}
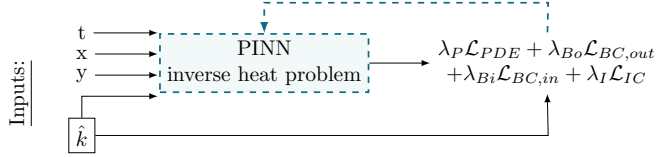
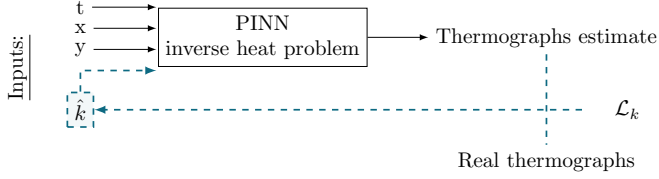
            \end{tikzpicture}
        }
        \caption{
            A PINN estimating the forward heat problem is iteratively optimized (\cref{fig:first}) with the estimated value $\hat{k}$ of $k$ (\cref{fig:second}).
        }
        \label{fig:all}
    \end{subfigure}
    \hfill
    \begin{subfigure}{\textwidth}
        \centering
        \scalebox{0.8}{\begin{tikzpicture}[node distance = 0.5cm, every node/.style={minimum height=0.5cm}]
    \node[align=center, rotate=90] (input) at (-3, 0) {\underline{Inputs:}};
    \node[align=center] (t) at (-2, 1) {t};
    \node[align=center] (x) at (-2, 0.65) {x};
    \node[align=center] (y) at (-2, 0.3) {y};
    \node[draw, align=center] (k)  at (-2, -0.7) {$\hat{k}$};

    \node[align=center, optimizedbox] (PINN) at (1, 0.5) {PINN\\inverse heat problem};

    \draw[-latex] (t) -- (-0.75, 1);
    \draw[-latex] (x) -- (-0.75, 0.65);
    \draw[-latex] (y) -- (-0.75, 0.3);
    \draw[-latex] (k) -- (-2, -0.05) -- (-0.75, -0.05);

    \node[align=center, right=1cm of PINN] (bc) {$\lambda_{P} \mathcal{L}_{PDE}   +   \lambda_{Bo} \mathcal{L}_{BC, out}$ \\ $   +   \lambda_{Bi} \mathcal{L}_{BC, in}   +   \lambda_{I} \mathcal{L}_{IC}$};
    \draw[-latex] (PINN) -- (bc);
    \draw[-latex] (k) -- (5.67, -0.7) -- (bc);
    \draw[-latex, optimized] (bc) -- (5.67, 1.5) -- (1, 1.5) -- (PINN);
\end{tikzpicture}}
        \caption{
            First step: a PINN is trained to estimate the forward heat problem based on a fixed value of $\hat{k}$, and the loss of the PDE $\mathcal{L}_{PDE}$, the boundary conditions in and out $\mathcal{L}_{BC,in}$ and $\mathcal{L}_{BC,out}$, and the initial condition $\mathcal{L}_{IC}$.
        }
        \label{fig:first}
    \end{subfigure}
    \hfill
    \begin{subfigure}{\textwidth}
        \centering
        \scalebox{0.8}{\begin{tikzpicture}[node distance = 0.5cm, every node/.style={minimum height=0.5cm}]
    \node[align=center, rotate=90] (input) at (-3, 0) {\underline{Inputs:}};
    \node[align=center] (t) at (-2, 1) {t};
    \node[align=center] (x) at (-2, 0.65) {x};
    \node[align=center] (y) at (-2, 0.3) {y};
    \node[align=center, optimizedbox] (k)  at (-2, -0.7) {$\hat{k}$};

    \node[draw, align=center] (PINN) at (1, 0.5) {PINN\\inverse heat problem};

    \draw[-latex] (t) -- (-0.75, 1);
    \draw[-latex] (x) -- (-0.75, 0.65);
    \draw[-latex] (y) -- (-0.75, 0.3);
    \draw[-latex, optimized] (k) --(-2, -0.05) -- (-0.75, -0.05);

    \node[align=center, right=1cm of PINN] (Te) {Thermographs estimate};
    \draw[-latex] (PINN) -- (Te);
    \node[align=center, below=1.5cm of Te] (T) {Real thermographs};
    \draw[-, optimized] (Te) -- (T);
    \draw[-latex, optimized] (6.25, -0.7) -- (k);

    \node[align=center] (loss) at (7, -0.7) {$\mathcal{L}_k$};
\end{tikzpicture}}
        \caption{
            Second step: $k$ is learned using the predicted temperature of the frozen PINN at the facade surface and measured thermographs as a loss to optimize $\hat{k}$.}
        \label{fig:second}
    \end{subfigure}

    \caption{PINN-it iteratively optimizes both a PINN and the estimation of the thermal conductivity $k$ in a 2-step process.
        In dashed lines are the optimized elements and losses at each step.
        First (\cref{fig:first}), a PINN is optimized to estimate the reverse heat problem given a fixed estimated $k$ value, i.e. $\hat{k}$.
        Then (\cref{fig:second}), $\hat{k}$ is optimized using a fixed PINN, using the difference between the estimated thermographs and measured ones as the loss.
        Steps 1 and 2 are iteratively repeated until convergence of $\hat{k}$.}
    \label{fig:intro:flowchart}
\end{figure}

In summary, there is a lack of methods able to estimate thermophysical properties under diverse environmental conditions, and with a simple, straightforward data collection approach.
To address those gaps, we propose a PINN-based methodology (PINN-it) for estimating the thermal conductivity $k$ of a material under varying environmental conditions.
The proposed approach involves a sequential optimization process: first, a PINN is trained to solve the forward-heat problem of a wall with a fixed thermal conductivity.
Subsequently, based on the hypothesis that a trained PINN allows for local optimization of the thermal conductivity toward its real value, an iterative optimization of both $k$ and the PINN is performed until convergence is achieved.
The methodology is designed to handle real-world observational variability, offering a robust framework for non-invasive thermal conductance estimations.
See \cref{fig:intro:flowchart} for a flowchart of the method.

PINN-it has been evaluated in terms of Mean Absolute Error (MAE) between the predicted and real thermal conductivity $k$.
We conduct the evaluations using real environmental data collected in situ from a Swiss governmental weather station and simulated wall temperature measurements from a finite-volume method (FVM) software.
Our evaluation demonstrates that the proposed method yields accurate estimates of k, highlighting the potential of physics-informed neural networks (PINNs) for in situ inverse problems that incorporate environmental data.

\section{Results}\label{sec2}

\subsection{Method overview and experiment setup}

The PINN-it framework---illustrated in \cref{fig:intro:flowchart}---iteratively estimates the thermal conductivity $k$ of a material through two alternating steps until convergence of $k$:
\begin{enumerate}
    \item Forward heat problem learning: A physics-informed neural network (PINN) approximates the temperature distribution in the wall for a fixed estimated conductivity $\hat{k}$, using Neumann boundary conditions, known initial conditions, and the governing heat equation.
          Environmental measurements are incorporated into the boundary constraints.
    \item Conductivity optimization: With the PINN weights frozen, $\hat{k}$ is optimized to minimize the discrepancy between real thermographic measurements and PINN-generated thermographs.
\end{enumerate}

The core hypothesis is that alternating between PINN retraining (for a given $\hat{k}$) and local optimization of $\hat{k}$ (for a fixed PINN) will converge to the true material conductivity.
A detailed formulation of the framework is provided in the Methods.

We conducted experiments on three homogeneous single leaf walls with fixed thermal conductivities of $k=0.75$, $k=2$, and $k=5$ [$W/m K$].
The remaining wall properties---density ($\rho$ [kg/m$^{3}$] = $2300$), specific heat capacity ($C_{p}$ [J/kgK] = $750$), and thickness ($0.3\,$m)---were selected to reflect typical concrete characteristics, using tabulated reference values.
To obtain realistic climatic conditions, we sourced meteorological data from the NABEL network\footnote{\url{https://www.empa.ch/web/s503/ambient-air-pollution}} (FOEN and Empa), recorded at a weather station in Lausanne, Switzerland.
This ensured similarity to in-situ building conditions (see Methods for detailed dataset and data generation specifications).
Ground-truth data was generated using an open-source finite volume method (FVM) solver.

To assess PINN-it under varying data availability, we generated synthetic thermographs via finite-element simulations under two sampling protocols:
\begin{enumerate}
    \item T$_{4-18}$: Thermographs captured every 15 minutes over a 4.5-hour simulation (18 total, with the first recorded at t = 15 min).
    \item T$_{1-5}$: Thermographs captured every 15 minutes during the final hour of the 4.5 hours simulation (5 thermographs total), modeling a short measurement window.
\end{enumerate}
Experiments were performed using 24 days (the first day of each month with sufficient data to be able to run the FVM software) spanning 2010-2011 and 2023-2024, ensuring coverage of all seasons in our tests.

\subsection{Inverse Problem with Steady State Condition}
\label{sec:inverse-solution}

\begin{table}[t]
    \centering
    \caption{This table shows the mean and median estimated conductivity and the MAE between PINN-it's predictions $\hat{k}$ and the actual values of $k$, assuming the wall temperature profile begins at steady state.
        Confidence interval (CI) calculated using bootstrapping by taking 1000 samples 10,000 times.
    }
    \label{tab:results-ss-seasonal}
    \begin{tabular}{llccccc}\toprule
$k$  & Season & mean & median & MAE [W/mK] & 95\% Conf Int & failed\\ \midrule
\multicolumn{7}{c}{T$_{4-18}$} \\ \midrule
\fs \multirow{5}{*}{0.75}  &  Year & 0.7893$\pm 0.0690$ & $0.7701$ & $0.0462$  & $0.0424 - 0.0501$ & 1 \\
 & Spring & $0.8290\pm 0.1082$ & $0.7841$ & $0.0790$  & $0.0730 - 0.0849$ & - \\
 & Summer & $0.7670\pm 0.0193$ & $0.7650$ & $0.0207$  & $0.0198 - 0.0215$ & - \\
 & Fall & $0.7866\pm 0.0442$ & $0.7895$ & $0.0422$  & $0.0400 - 0.0444$ & - \\
 & Winter & $0.7847\pm 0.0966$ & $0.7527$ & $0.0547$  & $0.0501 - 0.0595$ & - \\
 \midrule
\fs \multirow{5}{*}{2.0}  &  Year & 2.0083$\pm 0.0298$ & $2.0098$ & $0.0261$  & $0.0251 - 0.0271$ & 1 \\
 & Spring & $2.0113\pm 0.0355$ & $2.0183$ & $0.0309$  & $0.0300 - 0.0317$ & - \\
 & Summer & $2.0067\pm 0.0354$ & $2.0073$ & $0.0288$  & $0.0277 - 0.0298$ & - \\
 & Fall & $2.0133\pm 0.0332$ & $2.0265$ & $0.0311$  & $0.0303 - 0.0319$ & - \\
 & Winter & $2.0005\pm 0.0137$ & $2.0016$ & $0.0106$  & $0.0102 - 0.0110$ & - \\
 \midrule
\fs \multirow{5}{*}{5.0}  &  Year & 4.9977$\pm 0.1624$ & $5.0086$ & $0.0841$  & $0.0761 - 0.0926$ & 1 \\
 & Spring & $4.9679\pm 0.3340$ & $5.0585$ & $0.2084$  & $0.1951 - 0.2220$ & - \\
 & Summer & $5.0324\pm 0.0491$ & $5.0205$ & $0.0419$  & $0.0397 - 0.0441$ & - \\
 & Fall & $4.9906\pm 0.0973$ & $5.0086$ & $0.0651$  & $0.0613 - 0.0690$ & - \\
 & Winter & $4.9941\pm 0.0043$ & $4.9945$ & $0.0059$  & $0.0057 - 0.0061$ & - \\
 \midrule
\multicolumn{7}{c}{T$_{1-5}$} \\ \midrule
\fs \multirow{5}{*}{0.75}  &  Year & 2.1772$\pm 5.4951$ & $0.7754$ & $1.4328$  & $1.1099 - 1.7749$ & 1 \\
 & Spring & $0.7605\pm 0.0394$ & $0.7582$ & $0.0265$  & $0.0250 - 0.0281$ & - \\
 & Summer & $1.6419\pm 2.3103$ & $0.7754$ & $0.8942$  & $0.7663 - 1.0294$ & - \\
 & Fall & $0.8233\pm 0.1028$ & $0.7967$ & $0.0782$  & $0.0726 - 0.0839$ & - \\
 & Winter & $7.2540\pm 12.9647$ & $0.7793$ & $6.5040$  & $5.8041 - 7.2296$ & - \\
 \midrule
\fs \multirow{5}{*}{2.0}  &  Year & 3.3000$\pm 4.2789$ & $2.0453$ & $1.3015$  & $1.0464 - 1.5708$ & 1 \\
 & Spring & $4.0044\pm 4.3999$ & $2.0493$ & $2.0044$  & $1.7581 - 2.2507$ & - \\
 & Summer & $5.0191\pm 7.3405$ & $2.0227$ & $3.0191$  & $2.6110 - 3.4387$ & - \\
 & Fall & $2.0508\pm 0.0377$ & $2.0720$ & $0.0540$  & $0.0522 - 0.0559$ & - \\
 & Winter & $2.0268\pm 0.0304$ & $2.0256$ & $0.0297$  & $0.0282 - 0.0311$ & - \\
 \midrule
\fs \multirow{5}{*}{5.0}  &  Year & 6.5999$\pm 3.9855$ & $5.0599$ & $1.6165$  & $1.3780 - 1.8615$ & 0 \\
 & Spring & $5.0053\pm 0.0556$ & $5.0019$ & $0.0434$  & $0.0418 - 0.0450$ & - \\
 & Summer & $6.2775\pm 3.3066$ & $5.0269$ & $1.2915$  & $1.1092 - 1.4846$ & - \\
 & Fall & $5.4657\pm 1.0341$ & $5.1030$ & $0.4815$  & $0.4230 - 0.5405$ & - \\
 & Winter & $10.2337\pm 7.1560$ & $5.0625$ & $5.2337$  & $4.8431 - 5.6365$ & - \\\bottomrule\end{tabular}

\end{table}

\begin{figure}[t]
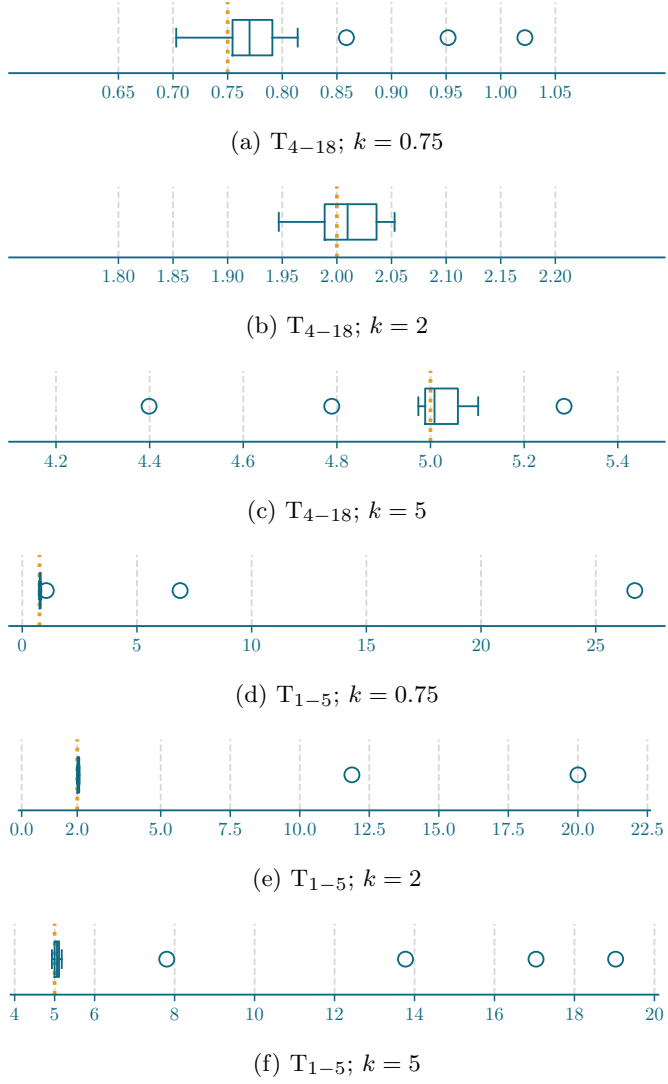

    \centering
    \begin{subfigure}[b]{\textwidth}
        \centering
        \scalebox{0.7}{\input{data/many/075/075_ss_k_boxplot.pgf}}
        \caption{T$_{4-18}$; $k=0.75$}
        \label{fig:ss_many_075}
    \end{subfigure}
    \hfill
    \centering
    \begin{subfigure}[b]{\textwidth}
        \centering
        \scalebox{0.7}{\input{data/many/200/200_ss_k_boxplot.pgf}}
        \caption{T$_{4-18}$; $k=2$}
        \label{fig:ss_many_200}
    \end{subfigure}
    \hfill
    \centering
    \begin{subfigure}[b]{\textwidth}
        \centering
        \scalebox{0.7}{\input{data/many/500/500_ss_k_boxplot.pgf}}
        \caption{T$_{4-18}$; $k=5$}
        \label{fig:ss_many_500}
    \end{subfigure}
    \hfill
    \begin{subfigure}[b]{\textwidth}
        \centering
        \scalebox{0.7}{\input{data/few/075/075_ss_k_boxplot.pgf}}
        \caption{T$_{1-5}$; $k=0.75$}
        \label{fig:ss_few_075}
    \end{subfigure}
    \hfill
    \begin{subfigure}[b]{\textwidth}
        \centering
        \scalebox{0.7}{\input{data/few/200/200_ss_k_boxplot.pgf}}
        \caption{T$_{1-5}$; $k=2$}
        \label{fig:ss_few_200}
    \end{subfigure}
    \hfill
    \begin{subfigure}[b]{\textwidth}
        \centering
        \scalebox{0.7}{\input{data/few/500/500_ss_k_boxplot.pgf}}
        \caption{T$_{1-5}$; $k=5$}
        \label{fig:ss_few_500}
    \end{subfigure}
    \caption{
        Boxplot distribution of the estimation of $k$ for all simulations with enforced steady-state condition at $t=0$.
    }
    \label{fig:boxplot_ss_t418}
\end{figure}

First, we evaluated the ability of PINN-it to estimate the thermal conductivity $k$ under the assumption that the wall reaches steady state at dawn (see Methods~\ref{sec:initial-condition} for details of the method's assumption).
The objective is to evaluate our method under similar conditions and assumptions to in-situ material property estimation campaigns; environmental conditions are uncontrolled, and the steady-state assumption reflects common practice of in-situ measurement campaigns.
To quantify accuracy, we computed the mean absolute error (MAE) between the estimated $\hat{k}$ and ground-truth $k$ values, along with its associated $95\%$ confidence interval (CI)---given the limited number of samples, the confidence intervals were obtained using bootstrapping with resampling by taking 1000 samples 10,000 times.
\cref{tab:results-ss-seasonal} shows the metric for $k=0.75$, $k=2$, and $k=5$ under the T$_{4-18}$ and T$_{1-5}$ scenarios.

Across all tested conditions---varying seasons, environmental factors, and sampling strategies---PINN-it consistently converged to the correct $k$ value, with only one failure case (1 December 2023) where the method failed to converge (apart from T$_{1-5}$ and $k=5.0$ where all days converged).
Since the method didn't converge to a specific value in this case, we excluded the failed runs from further analysis and explicitly state the number of failed convergences in the tables.
These results demonstrate that the PINN can learn to take into account environmental conditions and that our local optimization scheme for $k$---where the PINN locally approximates the forward heat problem for a given $k$, followed by optimization of $k$ with a fixed PINN until convergence---is effective.

\textbf{Performance under long sampling time:}
Under the T$_{4-18}$ protocol (18 thermographs over 4.5 hours), the distribution of estimates (\cref{fig:boxplot_ss_t418}) remained tightly centered around ground truth, showing robustness of the $k$ estimation under extended observation windows.
By season analysis (\cref{tab:results-ss-seasonal}) further validated consistency, with errors remaining stable across all conditions, demonstrating PINN-it's robustness to varying environmental condition variations and material properties.
Indeed, PINN-it achieved overall high precision: MAE $=0.0462$ [W/mK] ($95\%$ CI $0.0424 - 0.0501$) for $k=0.75$, MAE $=0.0261$ [W/mK] $95\%$ CI $0.0251 - 0.0271$) for $k=2$, and MAE $=0.0.0841$ [W/mK] ($95\%$ CI $0.0761 - 0.0926$) for $k=5$.

\textbf{Performance under limited observations:}
Under the T$_{1-5}$ sampling protocol (5 thermographs in the final hour), PINN-it is also able to accurately predict thermal conductivity $k$, with medians and distribution statistics comparable to the longer sampling protocol T$_{4-18}$.
However, while medians of the predicted $\hat{k}$ are the same as for the T$_{4-18}$ scenario, few outliers in the estimated values (2-4 per $k$ value) introduced variability in the MAE and mean (see the boxplot of \cref{fig:ss_few_075,fig:ss_few_200,fig:ss_few_500} compared to \cref{fig:ss_many_075,fig:ss_many_200,fig:ss_many_500})---the MAE increased to $1.4328$ [W/mK] ($95\%$ CI $1.1099 - 1.7749$) for $k=0.75$, $1.3015$ [W/mK] $95\%$ CI $1.0464 - 1.5708$) for $k=2$, and $1.6165$ [W/mK] ($95\%$ CI $1.3780 - 1.8615$) for $k=5$.
Notably, as seen in \cref{fig:ss_few_075,fig:ss_few_200,fig:ss_few_500}, incorrect estimates did not converge to a single erroneous value, suggesting that two independent measurement campaigns are sufficient for validation when data is sparse.
This makes PINN-it a valuable tool for predicting material properties, as the correct k can be reliably identified even when data collection cannot be conducted for a long period of time, regardless of environmental conditions.
In contrast, T$_{4-18}$ (\cref{fig:ss_many_075,fig:ss_many_200,fig:ss_many_500}) eliminated outliers entirely, underscoring the benefits of higher sampling density for robustness.

\subsection{Inverse Problem without Steady State Condition}
\label{sec:inverse-solution-non-steady}

This section evaluates the sensitivity of the framework to the steady-state assumption.
While the steady-state at dawn assumption is common in the literature, environmental variability may invalidate this assumption in practice.
Thus, instead of the conventional dawn initialization, the FVM simulation was initiated at steady state three days before the day used for data generation.
The simulation is then run for three days---taken into account environmental conditions---allowing a more realistic temperature profile at $t=0$ on the day the PINN-it method is evaluated.
It should be noted that in this setup, while the $t=0$ temperature might not be at steady state in the following experiments, the PINN-it method still assumes that the temperature profile at $t=0$ is at steady state, possibly creating a discrepancy.
\cref{tab:results-nss-seasonal} shows the metric for $k=0.75$, $k=2$, and $k=5$ under the T$_{4-18}$ and T$_{1-5}$ scenarios.

\begin{figure}[t]
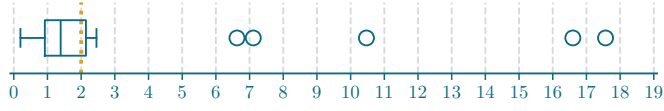
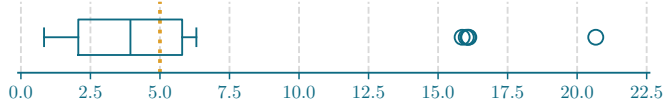
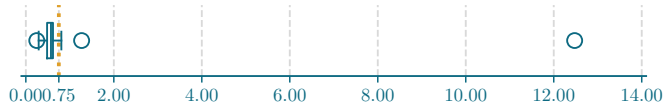
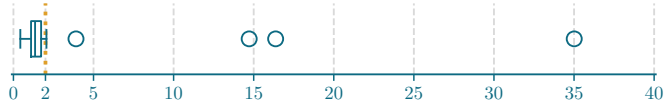
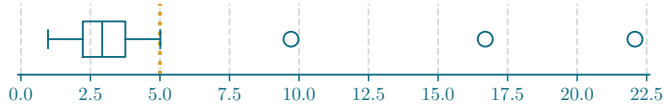

    \centering
    \begin{subfigure}[b]{\textwidth}
        \centering
        \scalebox{0.7}{\input{data/many/075/075_nonss_k_boxplot.pgf}}
        \caption{T$_{4-18}$; $k=0.75$}
        \label{fig:nonss:boxplot075:t418}
    \end{subfigure}
    \hfill
    \begin{subfigure}[b]{\textwidth}
        \centering
        \scalebox{0.7}{\input{data/many/200/200_nonss_k_boxplot.pgf}}
        \caption{T$_{4-18}$; $k=2$}
        \label{fig:nonss:boxplot200:t418}
    \end{subfigure}
    \hfill
    \begin{subfigure}[b]{\textwidth}
        \centering
        \scalebox{0.7}{\input{data/many/500/500_nonss_k_boxplot.pgf}}
        \caption{T$_{4-18}$; $k=5$}
        \label{fig:nonss:boxplot500:t418}
    \end{subfigure}
    \begin{subfigure}[b]{\textwidth}
        \centering
        \scalebox{0.7}{\input{data/few/075/075_nonss_k_boxplot.pgf}}
        \caption{T$_{1-5}$; $k=0.75$}
        \label{fig:nonss:boxplot075:t15}
    \end{subfigure}
    \hfill
    \begin{subfigure}[b]{\textwidth}
        \centering
        \scalebox{0.7}{\input{data/few/200/200_nonss_k_boxplot.pgf}}
        \caption{T$_{1-5}$; $k=2$}
        \label{fig:nonss:boxplot200:t15}
    \end{subfigure}
    \hfill
    \begin{subfigure}[b]{\textwidth}
        \centering
        \scalebox{0.7}{\input{data/few/500/500_nonss_k_boxplot.pgf}}
        \caption{T$_{1-5}$; $k=5$}
        \label{fig:nonss:boxplot500:t15}
    \end{subfigure}
    \caption{
        Boxplot distribution of the estimation of $k$ for all simulations without steady-state condition.
        While the estimation of $k$ is less stable than with guaranteed steady-state condition, PINN-it mostly converges toward the correct value of $k$.
    }
    \label{fig:nonss:boxplot}
\end{figure}

While, due to the deviations from the steady-state assumption, the accuracy of the estimated $k$ values decreased compared to when the steady-state assumption is respected, most $k$ estimations still converged toward the correct solution.
Looking at the distribution of the estimations (\cref{fig:nonss:boxplot}), one can see that uncertainty in the estimation is mostly due to a few large outliers, with most estimations close to the real value of $k$.
These results demonstrate that, while near-steady-state initial conditions enhance the performance of PINN-it as an effective surrogate model for the direct problem---ensuring robust k estimates for field applications---PINN-it still exhibits some adaptability to arbitrary environmental conditions, even under relaxed assumptions, unlike existing approaches, which rely on steady-state conditions and/or ignore environmental conditions.

\begin{figure}[t]
    \centering
    \input{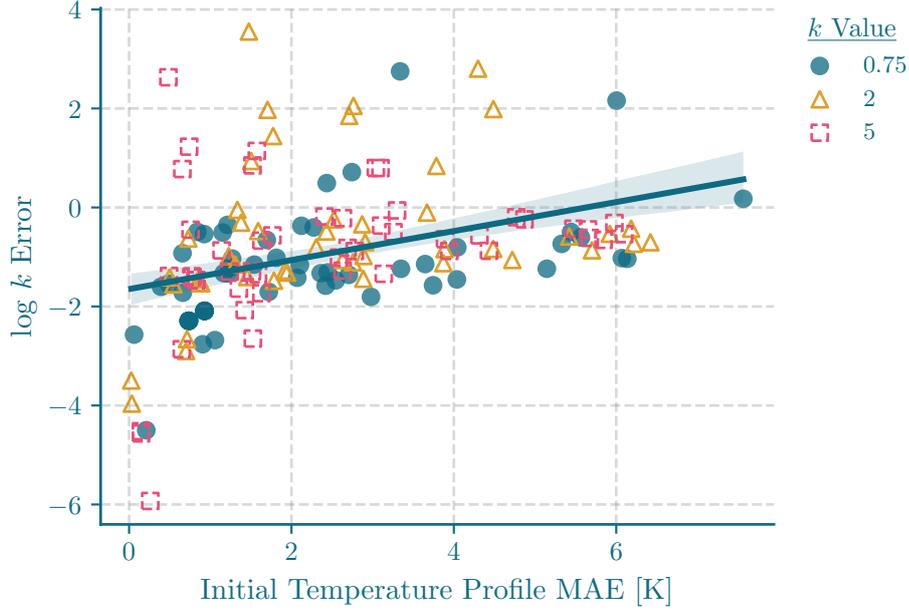}
    \caption{
        Initial temperature profile MAE vs log absolute $k$ error for both T$_{4-18}$ and T$_{1-5}$.
        The initial temperature profile MAE is calculated as the MAE between the temperature profile at $t=0$ from the openFOAM simulations and the temperature profile if the wall were in steady state at $t=0$.
        Log $k$ error compares the error in the predicted value of $k$ from PINN-it and the value of $k$ used to run the openFOAM simulation.
        While higher errors in predictions are associated with higher MAE values, PINN-it is still able to estimate $k$ with a large initial temperature profile's MAE.
    }
    \label{fig:wall-profile-mae-relative-k-error-many}
\end{figure}

Interestingly, the larger relative error of this experiment highlights a discrepancy in initial wall temperature profiles ($t = 0$) between 3-day spin-up and true steady-state simulations---which is a common assumption for in-situ measurements in today's practice~\cite{noauthor_iso_nodate}.
In \cref{fig:wall-profile-mae-relative-k-error-many}, we correlate this deviation (i.e., the error between FVM simulation profile at $t=0$ after three day spin-up period and the temperature profile if steady state was reached) with $\hat{k}$ estimation errors.
While higher initial-profile MAEs (vs. steady state) correspond to larger $\hat{k}$ errors, the spread of relative errors is still broad, suggesting that compounding environmental factors may amplify estimation errors under non-steady-state conditions.
Critically, PINN-it retains functional estimation capability even at high initial-profile MAEs, albeit with reduced precision.

\begin{table}[t]
    \caption{
        This table shows the mean and median estimated conductivity and the MAE between PINN-it's predictions $\hat{k}$ and the actual values of $k$, without assuming the wall temperature profile begins at steady state.
        Confidence interval (CI) calculated using bootstrapping by taking 1000 samples 10,000 times. }
    \label{tab:results-nss-seasonal}
    \begin{tabular}{llccccc}\toprule
$k$  & Season & mean & median & MAE [W/mK] & 95\% Conf Int & failed\\ \midrule
\multicolumn{7}{c}{T$_{4-18}$} \\ \midrule
\fs \multirow{5}{*}{0.75}  &  Year & 0.9726$\pm 1.4329$ & $0.5495$ & $0.6238$  & $0.5458 - 0.7037$ & 1 \\
 & Spring & $1.8026\pm 3.0523$ & $0.5276$ & $1.5484$  & $1.3935 - 1.7042$ & - \\
 & Summer & $0.8325\pm 0.6790$ & $0.5109$ & $0.5238$  & $0.5021 - 0.5464$ & - \\
 & Fall & $0.8016\pm 0.6552$ & $0.5513$ & $0.3863$  & $0.3571 - 0.4160$ & - \\
 & Winter & $0.5783\pm 0.1157$ & $0.5989$ & $0.1717$  & $0.1652 - 0.1780$ & - \\
 \midrule
\fs \multirow{5}{*}{2.0}  &  Year & 3.4028$\pm 4.8637$ & $1.3923$ & $2.6692$  & $2.4200 - 2.9448$ & 1 \\
 & Spring & $4.3637\pm 6.8541$ & $1.4175$ & $3.4750$  & $3.1260 - 3.8232$ & - \\
 & Summer & $1.7565\pm 2.3916$ & $0.8870$ & $1.7035$  & $1.6146 - 1.7961$ & - \\
 & Fall & $3.4757\pm 3.6745$ & $1.5320$ & $2.3960$  & $2.2233 - 2.5770$ & - \\
 & Winter & $4.6444\pm 7.2370$ & $1.5767$ & $3.5978$  & $3.2356 - 3.9821$ & - \\
 \midrule
\fs \multirow{5}{*}{5.0}  &  Year & 6.0538$\pm 5.9530$ & $3.9382$ & $4.0851$  & $3.8239 - 4.3488$ & 1 \\
 & Spring & $7.7729\pm 6.6077$ & $4.7970$ & $4.6208$  & $4.3311 - 4.9142$ & - \\
 & Summer & $1.9371\pm 0.9663$ & $1.7653$ & $3.0629$  & $3.0074 - 3.1180$ & - \\
 & Fall & $5.5330\pm 4.8610$ & $3.8464$ & $3.1251$  & $2.9293 - 3.3388$ & - \\
 & Winter & $9.6602\pm 8.1239$ & $4.7148$ & $6.0128$  & $5.6324 - 6.3976$ & - \\
 \midrule
\multicolumn{7}{c}{T$_{1-5}$} \\ \midrule
\fs \multirow{5}{*}{0.75}  &  Year & 1.0517$\pm 2.4418$ & $0.5717$ & $0.7234$  & $0.5897 - 0.8739$ & 1 \\
 & Spring & $0.4553\pm 0.1832$ & $0.4372$ & $0.2947$  & $0.2845 - 0.3047$ & - \\
 & Summer & $2.2027\pm 4.5307$ & $0.5325$ & $1.8974$  & $1.6560 - 2.1417$ & - \\
 & Fall & $0.6895\pm 0.2601$ & $0.5955$ & $0.2085$  & $0.2003 - 0.2172$ & - \\
 & Winter & $0.5657\pm 0.2088$ & $0.5788$ & $0.2130$  & $0.2041 - 0.2222$ & - \\
 \midrule
\fs \multirow{5}{*}{2.0}  &  Year & 4.0845$\pm 7.9061$ & $1.3514$ & $3.3126$  & $2.8674 - 3.7815$ & 0 \\
 & Spring & $7.9198\pm 15.1557$ & $1.5102$ & $7.2847$  & $6.4838 - 8.0801$ & - \\
 & Summer & $3.7254\pm 6.1970$ & $1.2332$ & $3.0647$  & $2.7501 - 3.3719$ & - \\
 & Fall & $1.6946\pm 0.9931$ & $1.3514$ & $0.8500$  & $0.8218 - 0.8791$ & - \\
 & Winter & $4.0260\pm 6.0024$ & $1.5129$ & $3.0856$  & $2.7901 - 3.3852$ & - \\
 \midrule
\fs \multirow{5}{*}{5.0}  &  Year & 4.5946$\pm 5.0514$ & $2.9228$ & $3.3185$  & $3.0994 - 3.5533$ & 0 \\
 & Spring & $2.8294\pm 1.5132$ & $2.4420$ & $2.1706$  & $2.0859 - 2.2543$ & - \\
 & Summer & $5.2099\pm 7.4793$ & $2.4063$ & $4.6698$  & $4.3611 - 4.9866$ & - \\
 & Fall & $4.2686\pm 2.5781$ & $3.6021$ & $2.0814$  & $1.9955 - 2.1669$ & - \\
 & Winter & $6.2951\pm 6.9796$ & $3.3366$ & $4.5532$  & $4.2921 - 4.8178$ & - \\\bottomrule\end{tabular}

\end{table}

\textbf{Performance under long sampling time:}
For all tested values of $k$, 24 experiments out of 25 converged---with the exception of the 1st December 2023 for $k=0.75$, and the 5th of January for $k=2$ and $k=5$---as in \cref{sec:inverse-solution}.
The accuracy of the setimated parameter $\hat{k}$ decreased under non-steady state condition with MAE of $0.6238$ ($95\%$ CI $0.5458 - 0.7037$), $2.6692$ ($95\%$ CI $2.4200 - 2.9448$), and $4.0851$ ($95\%$ CI $3.8239 - 4.3488$) for each $k=0.75$, $k=2.0$, and $k=5.0$ respectively---compared to $0.0462$, $0.0261$, and $0.0841$ under a respected steady-state assumptions.
Despite higher MAE, the median estimated remained relatively close to ground truth (medians: $0.5495$ $1.3923$, and $3.9382$, for $k=0.75$, $k=2$, and $k=5$ respectively).
The distributions (see \cref{fig:nonss:boxplot075:t418,fig:nonss:boxplot200:t418,fig:nonss:boxplot500:t418}) reveal that large MAE stem from large isolated outliers,
suggesting robust central tendency despite sporadic high-magnitude errors.

\textbf{Performance under limited observations:}
Contrary to what was observed when the steady-state assumption was respected in \cref{sec:inverse-solution}, without the stead-state assumption the mean, median, and MAE values of $\hat{k}$ are similar (with only slightly larger standard deviation and MAE) to the ones found for T$_{4-18}$, highliting the robustness of PINN-it for low amount of measurements.
This slight disparity in standard deviation and MAE arises from some larger outliers in T$_{1-5}$, though the number of outliers remained consistent between scenarios and the difference between them is not significant---this is visible by looking at the distribution of the estimated $k$ values in \cref{fig:nonss:boxplot075:t15,fig:nonss:boxplot200:t15,fig:nonss:boxplot500:t15} compared to \cref{fig:nonss:boxplot075:t418,fig:nonss:boxplot200:t418,fig:nonss:boxplot500:t418} d to f, where the boxes for both scenarios are similar.
The persistence of similar metrics across sampling strategies (T$_{4-18}$ vs. T$_{1-5}$) indicates that PINN-it's accuracy and precision are preserved, with degraded performance limited to outlier sensitivity.

\subsection{PINN Forward Problem Accuracy}
\label{sec:direct-solution}

In this section, we assess the PINN's ability to solve the direct problem, thereby demonstrating its potential to act as a surrogate forward model for heat transfer in a building facade.
Evaluation involved training the PINN $\mathcal{U}$ (see Methods) until convergence and comparing the predictions of temperature throughout time and space to OpenFOAM FVM results.
Temperature values were compared between the simulation and PINN at intervals of 0.5mm within the wall and at every 5 minutes of simulation time.
The simulation was run for 4 hours and 30 minutes, resulting in 54 comparison measurements per day.
MAE between the OpenFOAM FVM simulations and PINN predictions was used to evaluate the accuracy of the estimated temperature profile.

\begin{table}[t]
    \centering
    \caption{Mean Root mean square error between PINN prediction of temperature and FVM simulation for all simulations by season, each group contains 6 simulations. Error was evaluated every 0.5 mm within the wall and every 5 minutes of simulation time. }
    \label{tab:results-direct-season}
    \begin{tabular}{llcc}\toprule
$k$& Season & MAE {[}W/mK{]} & 95\% Confidence Interval  \\ \midrule
\fs 0.75 & Year & $0.4194 \pm 0.5546 $ & $0.3860$ - $0.4535$ \\
0.75 & Spring & $0.4830 \pm 0.4214$  & $0.4575$ - $0.5091$ \\
0.75 & Summer & $0.3666 \pm 0.3385$  & $0.3457$ - $0.3875$ \\
0.75 & Fall & $0.3871 \pm 0.7817$  & $0.3395$ - $0.4374$ \\
0.75 & Winter & $0.4698 \pm 0.5236$  & $0.4386$ - $0.5025$ \\
 \midrule
\fs 2.0 & Year & $0.0129 \pm 0.0130 $ & $0.0121$ - $0.0137$ \\
2.0 & Spring & $0.0223 \pm 0.0183$  & $0.0212$ - $0.0234$ \\
2.0 & Summer & $0.0109 \pm 0.0071$  & $0.0105$ - $0.0113$ \\
2.0 & Fall & $0.0131 \pm 0.0137$  & $0.0123$ - $0.0140$ \\
2.0 & Winter & $0.0065 \pm 0.0036$  & $0.0063$ - $0.0067$ \\
 \midrule
\fs 5.0 & Year & $0.0073 \pm 0.0067 $ & $0.0069$ - $0.0077$ \\
5.0 & Spring & $0.0118 \pm 0.0115$  & $0.0111$ - $0.0125$ \\
5.0 & Summer & $0.0069 \pm 0.0030$  & $0.0067$ - $0.0071$ \\
5.0 & Fall & $0.0067 \pm 0.0044$  & $0.0064$ - $0.0070$ \\
5.0 & Winter & $0.0037 \pm 0.0014$  & $0.0036$ - $0.0038$ \\\bottomrule\end{tabular}

\end{table}

The results of this experiment show that the PINN is able to act as a solution to the heat diffusion equation, demonstrating an MAE under $0.5$ K across seasons and $k$ values.
This suggests that the PINN is likely to be effective in estimating the parameter $k$ when using the PINN to predict surface temperatures for thermograph comparison.

\cref{tab:results-direct-season} presents the MAE in predicted temperature (yearly and seasonally) between the FVM and PINN simulations, as well as the confidence intervals associated with the MAE.
Across all simulations, the PINN achieves a mean MAE of $0.4194 \pm 0.5546$~[K], with a 95\% CI of $[0.3860, 0.4535]$ for $k = 0.75$, $0.0129 \pm 0.0130$ with a 95\% CI of $[0.0121, 0.0137]$ for $k = 2$, and $0.0073 \pm 0.0067$ with a 95\% CI of $[0.0069, 0.0077]$ for $k = 5$ highlighting the PINN's capability to predict the wall's temperature profile with high accuracy.
The temperature predictions of the PINN were accurate for both wall setups, with higher accuracy for the wall with $k$ = 2.

Furthermore, to understand $\mathcal{U}$'s stability under varying environmental conditions, \cref{tab:results-direct-season} provides the MAE of $\mathcal{U}$ for each season individually.
Since, for a given value of $k$, the metrics are stable between seasons, it can be concluded that $\mathcal{U}$ is able to predict wall temperature accurately across seasons.

\section{Discussion}\label{sec12}

Physics-informed neural networks (PINNs) have demonstrated promise in solving complex problems by approximating solutions to partial differential equations.
However, their reliance on controlled environments and known material properties limits their applicability to inverse problems, such as the inverse heat problem.
In contrast, PINN-it showcases a novel approach to determining thermophysical properties while accounting for real-world environmental data.
Unlike previous methods, PINN-it converges to accurate material properties without requiring controlled conditions.
This is achieved by sequentially training a PINN to learn the direct heat problem with fixed thermophysical parameters and known environmental conditions, followed by optimizing thermal conductivity $k$ using a frozen PINN and measured thermographs.

Experimental results confirm that the proposed methodology accurately predicts thermal conductivity when the initial condition of the wall in the PINN simulation is at steady-state.
The PINN's ability to take into account environmental conditions and PINN-it's robustness across seasons and environmental factors underscores its effectiveness as a non-invasive estimation tool.
This work highlights the potential of PINNs for precise material property modeling, even with measured environmental data and without a controlled environment.

Interestingly, experiments in \cref{sec:inverse-solution-non-steady} revealed that, when initial conditions were simulated rather than assumed to be at steady-state, the facade did not reach equilibrium---contrary to common assumptions in the state-of-the-art and in situ campaigns.
While PINN-it exhibited sensitivity to initial conditions, the overall accuracy of the $k$ estimation remained notable; improving the initial condition estimation based on environmental data is a key focus for future work.
Additionally, PINN-it was evaluated on a simple, single-material wall, whereas real-world structures often feature multiple materials.
Future iterations of the framework will extend to multi-material walls.

In conclusion, this study establishes a foundational framework for applying physics-informed neural networks (PINNs) to heat transfer analysis and thermophysical property retrieval.
Our results demonstrate PINNs' capacity to resolve ill-posed inverse problems, going beyond controlled laboratory environments.
The methodology holds promise for broader applications where in situ material property measurement remains challenging.
As machine learning applications in this domain remain limited, our work paves the way for further advancements in data-driven material property estimation.
Beyond academic implications, PINN-it offers practical advantages by reducing measurement time from days to hours, cutting retrofit costs, and delivering robust estimates with as few as five thermographs, minimizing hardware dependencies and cost.

\section{Methods}\label{sec11}

The PINN-it method aims to estimate the thermal conductivity $k$ of a wall to calculate the thermal resistance $R$ of buildings' facades:
\begin{equation}
    R = \frac{b}{k}
\end{equation}
where $b$ is the wall thickness.

In PINN-it, a PINN---referred to as $\mathcal{U}$---acts as a solution to the heat diffusion equation of the wall---described in \cref{sec:method:subsec:diffusion}---at any given location and time for a particular value of $k$.
Once the PINN is trained, it can predict surface temperatures of the wall and be compared to external temperature readings, allowing for the optimization of $k$ by performing gradient descent on the error between the external temperature readings and the predicted temperature from the PINN.
Estimating $k$ involves iteratively training the PINN for a particular value of $k$ until it adequately represents a solution to the heat diffusion equation, and then optimizing $k$ to minimize the thermographic loss $\mathcal{L}_{TC}$ based on the external temperature readings.
The full algorithm is detailed in \cref{alg:train} and a flowchart of the method is shown in \cref{fig:intro:flowchart}.

\begin{algorithm}[t]
    \KwOut{$\hat{k}$ the optimized thermal conductivity}
    Initialize PINN train state\;
    $\hat{k}_{0} \leftarrow $ mid point of looked up table values\;
    \For{$n \leftarrow 1$ \KwTo $training\_steps$}{
        \While{$\mathcal{L}_{PDE} > \mathcal{L}_{PDE,\:min}$ {\upshape \textbf{and}} $\mathcal{L}_{BC,   out} > \mathcal{L}_{BC,\:min}$ {\upshape \textbf{and}} $\mathcal{L}_{BC,   in} > \mathcal{L}_{BC,\:min}$ {\upshape \textbf{and}} $\mathcal{L}_{IC} > \mathcal{L}_{IC,\:min}$}{
            Sample collation points in the physical and temporal domains\;
            Obtain collation points in $k$ domain by drawing from a truncated normal distribution centered at $\hat{k}_{n}$\ with limits of $\hat{k}_{min}$ and $\hat{k}_max$ taken from table values\;
            $\mathcal{L}_{total} \leftarrow$ $\lambda_{PDE}\mathcal{L}_{PDE}$, $\lambda_{BC,\:out}\mathcal{L}_{BC,\:out}$, $\lambda_{BC,\:in}\mathcal{L}_{BC,\:in}$,  and $\lambda_{IC}\mathcal{L}_{IC}$\; %
            Update PINN parameters\;}

        \For{$m \leftarrow 1$ \KwTo nb thermographs}{
            Estimate the surface temperature $T$ at capture time of thermograph $T_m$\;
            $\mathcal{L}_{k} \leftarrow \mathcal{L}_{k} + (||T - T_m||)$
        }
        $\hat{k}_{n+1} \leftarrow $ Update $\hat{k}_{n}$ based on $\mathcal{L}_{k}$\;
        Reset PINN optimizer;

    }
    \Return $\hat{k}$\;

    \caption{Algorithm for obtaining $k$ using thermograph and PINN methodology. $\mathcal{L}_{PDE,\:min}$, $\mathcal{L}_{BC,\:min}$, $\mathcal{L}_{IC,\:min}$ are the minimum values of losses from $\mathcal{U}$ before optimizing $\hat{k}$}
    \label{alg:train}
\end{algorithm}

\subsection{Heat Diffusion Problem}
\label{sec:method:subsec:diffusion}

As PINNs are used to solve differential equations, the appropriate heat diffusion equation, boundary conditions, and initial conditions need to be chosen for the case of a 1-dimensional homogeneous single-leaf wall undergoing forced convective heat transfer and solar irradiance on the outside surface, as well as natural convection on the inside surface.
The wall is assumed to have a uniform thickness and constant thermal properties.
Thus, the generalized heat diffusion equation can be written as:
\begin{equation}
    \label{eq:PDE-dimensioned}
    \frac{\partial T}{\partial t} = \frac{k}{C_{p}   \rho}   \frac{\partial^{2} T}{\partial x^{2}}
\end{equation}
Where $T$ represents temperature, $t$ time, $k$ is the thermal conductivity, $C_{p}$ the specific heat capacity, $\rho$ the density, and $x$ the distance within the wall.
To solve this partial differential equation, either numerically or through the use of a PINN, boundary and initial conditions must be provided.

\subsubsection{Boundary conditions}

Neumann boundary conditions, which describe the rate of heat flow at a boundary, can be used for cases of convective heat transfer.
The Neumann boundary condition for convective heat transfer to a flat surface is expressed as:
\begin{equation}
    \label{eq:neumann-general}
    k \frac{\partial T}{\partial x} = h\left(T_{wall} - T_{\infty}  \right),
\end{equation}
with $T_{\infty}$ the ambient temperature and $T_{wall}$ the temperature at the surface.

Given the expression of a Neumann boundary condition, we can obtain two boundary conditions---one on the inside and the other on the outside of the wall---for our PDE by substituting the appropriate values for $h$, $T_{wall}$, and $T_{\infty}$.
Notably, the heat transfer on the outdoor surface of the wall also involves solar irradiation, which is added to the right side of equation \cref{eq:neumann-general}.

Therefore, the heat flux on the outside surface of the wall can be described as:
\begin{equation}
    \label{eq:BC-out}
    k \frac{\partial T}{\partial x}\bigg\vert_{x=0} = h_{out}\left(T_{wall}\vert_{x=0} - T_{\infty,\:out}  \right) + (1-\alpha) Q_{dir} + Q_{diff} ,
\end{equation}
where $h_{out}$ is the outdoor convective heat transfer coefficient, $T\vert_{x=0}$ is the temperature on the outside surface of the wall,  $T_{\infty,\:out}$ is the outdoor ambient air temperature, $\alpha$ is the wall surface albedo, $Q_{dir}$ is the portion of solar irradiance that is direct and $Q_{diff}$ is the portion of solar irradiance that is diffuse.
\cref{eq:BC-out} can be simplified by incorporating the sol-air temperature, which represents the outdoor air temperature necessary to achieve the same heat transfer as the combined effect of convective heat flux and solar irradiation:
\begin{equation}
    \label{eq:solair}
    \  T_{sol,air} \;=\; T_{\infty,\:out}\;+\;\frac{(1-\alpha) Q_{dir} + Q_{diff}}{h_{out}}\
\end{equation}
By substituting $T_{\infty, \text{out}}$ with the sol-air temperature expression, \cref{eq:BC-out} simplifies to:
\begin{equation}
    \label{eq:BC-out-solair}
    k \frac{\partial T}{\partial x}\vert_{x=0}  = h_{out}\left(T_{wall}\vert_{x=0} - T_{sol,air}  \right)
\end{equation}

On the other hand, the heat flux on the indoor surface of the wall undergoes natural convection and can be modelled using the measured indoor air temperature:
\begin{equation}
    \label{eq:BC-in}
    k \frac{\partial T}{\partial x}\vert_{x=b}  = h_{in}\left(T_{\infty,\:in} - T_{wall}\vert_{x=b} \right)
\end{equation}
where  $T_{wall}\vert_{x=b}$ is the temperature on the inside surface of the wall, $T_{\infty,\:in}$ is the indoor air temperature, and $h_{in}$ is the indoor heat transfer coefficient which is a function of $T_{wall}\vert_{x=b}$ and $T_{\infty,\:in}$.

\subsubsection{Initial condition}
\label{sec:initial-condition}
To solve the partial differential equation (PDE) associated with heat transfer, the initial temperature profile is needed.
However, the temperature profile of a building facade cannot be measured, and an assumption must be made to approximate an initial temperature profile.
Assuming that, over the course of nighttime, the building facade approximately reaches steady state with its surroundings, the initial condition can be estimated by solving the diffusion equation at steady state using the outdoor air temperature at dawn, wind speed, indoor air temperature, and an approximation of the indoor heat transfer coefficient:
\begin{equation}
    \label{eq:SS}
    \ 0 = \frac{k}{C_{p}\rho}   \frac{\partial^{2} T}{\partial x^{2}} \
\end{equation}
With $t=0$ representing the time at sunrise.
From solving \cref{eq:SS} it can be shown that the initial condition $T_{0}(x)$ will take the form:
\begin{equation}
    \label{eq:ic-dimensioned}
    T_{0} (x) \;=\; \frac{\left.T\right\vert_{x=b,\:t=0} - \left.T\right\vert_{x=0,\:t=0}}{b} x + \left.T\right\vert_{x=0,\:t=0}
\end{equation}
with $\left.T\right\vert_{x=b,\:t=0}$ and $\left.T\right\vert_{x=0,\:t=0}$ respectively represented as:
\begin{equation}
    \label{eq:inside-wall-temp-0}
    \ \left.T\right\vert_{x=b}\;=\; \frac{b\:h_{out}h_{in}T_{\infty,\:in} \;+\; h_{out}k\:T_{\infty,\:out} \;+\; h_{in}k\:T_{\infty,\:in}}{b\:h_{out}h_{in} \;+\; k(h_{out} \:+\: h_{in})} \
\end{equation}

\begin{equation}
    \label{eq:outside-wall-temp-0}
    \ \left.T\right\vert_{x=0}\;=\; \frac{b\:h_{out}h_{in}T_{\infty,out} \;+\; h_{out}k\:T_{\infty,\:out} \;+\; h_{in}k\:T_{\infty,\:in}}{b\:h_{out}h_{in} \;+\; k(h_{out} \:+\: h_{in})} \
\end{equation}

\subsection{PINN}
\label{sec:method:subsec:PINN}

To solve the heat diffusion problem of \cref{sec:method:subsec:diffusion} a PINN is trained to estimate the problem's solution given a set of known physical parameters.
The PINN is a function of the weights of the network $W$, time $t$, position $x$, and thermal conductivity $K$, and outputs a temperature $T$.
\begin{equation}
    \ \mathcal{U}(W \vert \:t,\:x,\:k)\;=\;T   \
\end{equation}
Non-dimensionalization (i.e., normalization) of the variables in the network has been shown to help with the lack of consistent network initialization schemes, mitigating the disparities in variable scales, and improving convergence~\cite{wang_experts_2023}.
Thus, all variables---time, distance, $K$, and temperature---are normalized to range from 0 to 1.
Dimensionless time $\tau$ and dimensionless position $\xi$ are expressed as:
\begin{align}
    \label{eq:tau}
    \tau & = \frac{t}{t_{total}} \\
    \label{eq:xi}
    \xi  & = \frac{x}{b}
\end{align}
with $t_{total}$, the total time of simulation (the time from sunrise to the time of the last available thermograph).
Dimensionless temperature $\Theta$ and dimensionless thermal conductivity $K$ are expressed as
\begin{align}
    \label{eq:Theta}
    \Theta & = \frac{T - T_{min}}{T_{max} - T_{min}} \\
    \label{eq:K}
    K      & = \frac{k - k_{min}}{k_{max} - k_{min}}
\end{align}
Where $T_{max}$ and $T_{min}$ are the maximum and minimum temperatures in the simulation, and $k_{max}$ and $k_{min}$ are the maximum and minimum expected values of k for the material of the wall, obtained from tabular values.
Thus, $\mathcal{U}$ is expressed as:
\begin{equation}
    \mathcal{U}(W \: \vert \: \tau, \: \xi, \: K) = \Theta
\end{equation}

The non-dimensionalization performed in equations \cref{eq:tau,eq:xi,eq:Theta,eq:K}, is applied to the heat diffusion problem and the boundary conditions of \cref{eq:PDE-dimensioned,eq:BC-out-solair,eq:BC-in,eq:ic-dimensioned}:
\begin{equation}
    \label{eq:PDE-nondimensioned}
    \frac{1}{t_{total}} \frac{\partial \Theta}{\partial \tau} = \frac{K}{C_{p} \rho b^{2}}  \frac{\partial^{2}\Theta}{\partial \xi^{2}}
\end{equation}
\begin{equation}
    \label{eq:BC-out-solair-nondimensioned}
    \frac{K}{b} \frac{\partial \Theta}{\partial \xi}\vert_{\xi=0}  = h_{out}\left(\Theta_{wall}\vert_{\xi=0} - \Theta_{sol,air}  \right)
\end{equation}
\begin{equation}
    \label{eq:BC-in-nondimensioned}
    \frac{K}{b} \frac{\partial \Theta}{\partial \xi}\vert_{\xi=1}  = h_{in}\left(\Theta_{\infty,\:in} - \Theta_{wall}\vert_{\xi=1} \right)
\end{equation}
\begin{equation}
    \label{eq:ic-nondimensioned}
    \Theta_{0} (x) = \left(\left.\Theta\right\vert_{\xi=1,\:\tau=0} - \left.\Theta\right\vert_{\xi=0,\:\tau=0}\right) \xi + \left.\Theta\right\vert_{\xi=0,\:\tau=0}
\end{equation}

Using the above equations, the loss functions of $\mathcal{U}$ can be established as the mean squared error of the residuals of the heat diffusion equation, boundary conditions (of the outside and inside surfaces), and initial condition, which are evaluated at $N$ collocation points.
Collocation points are points in time $\tau_{i}$, space $\xi_{i}$, and thermal conductivity $K_{i}$, where the PINN is evaluated to obtain an estimated temperature $\Theta_{i}$, and are sampled uniformly along the length of the wall.
\begin{equation}
    \label{eq:PDE-loss}
    \ \mathcal{L}_{PDE}  =  \frac{1}{N_{PDE}} \mathlarger{\sum}^{N_{PDE}}_{i=0}
    \left\vert
    \left\vert
    \frac{\partial \mathcal{U}}{\partial \tau}
    \bigg\vert_{\tau_{i},\xi_{i},K_{i}}  -  \frac{K_{i}\: t_{total}}{C_{p} \:  \rho  \: b^{2}}
    \frac{\partial^{2}\mathcal{U}}{\partial   \xi^{2}}
    \bigg\vert_{\tau_{i},\xi_{i},K_{i}}
    \right\vert
    \right\vert^{2} \
\end{equation}
\begin{equation}
    \label{eq:BC-loss-out}
    \ \mathcal{L}_{BC,\:out}  =   \frac{1}{N_{BC}} \mathlarger{\sum}^{N_{BC}}_{i=0} \left\vert \left\vert \left(\left.\mathcal{U}\right\vert_{\tau_{i},\xi=0,K_{i}}  -  \Theta_{sol,air} \right)  -  \frac{K_{i}}{b \: h_{out}}  \left.\frac{\partial \mathcal{U}}{\partial \xi}\right\vert_{\tau_{i},\xi=0,K_{i}} \right\vert \right\vert^{2} \
\end{equation}
\begin{equation}
    \label{eq:BC-loss-in}
    \ \mathcal{L}_{BC,\:in}  =  \frac{1}{N_{BC}} \mathlarger{\sum}^{N_{BC}}_{i=0} \left\vert \left\vert \left(\Theta_{\infty,\:in}  -   \left.\mathcal{U}\right\vert_{\tau_{i}\xi=1,K_{i}} \right)  -  \frac{K_{i}}{b\:h_{in}}  \left.\frac{\partial \mathcal{U}}{\partial \xi}\right\vert_{\tau_{i},\xi=1,K_{i}} \right\vert \right\vert^{2} \
\end{equation}
\begin{equation}
    \label{eq:IC-loss}
    \ \mathcal{L}_{IC}  =  \frac{1}{N_{IC}} \sum^{N_{IC}}_{i=0} \left\vert \left\vert \Theta_{0} - \left.\mathcal{U}\right\vert_{\tau=0} \right\vert \right\vert^{2} \
\end{equation}
The losses described above can exist at different magnitudes during the training of the PINN, leading to the potential for preferential training of one loss over another.
Thus, the self-adjusting L2 norm weighting scheme developed by \citet{wang_experts_2023} is used to balance the previously described losses---for more details on the weighting scheme, we refer the reader to the original paper.

Finally, the final loss term of the PINN is:
\begin{equation}
    \label{eq:total-loss}
    \ \mathcal{L}_{Total} \;=\; \lambda_{PDE} \mathcal{L}_{PDE}   +   \lambda_{BC,\:out} \mathcal{L}_{BC,\:out}   +   \lambda_{BC,\:in} \mathcal{L}_{BC,\:in}   +   \lambda_{IC} \mathcal{L}_{IC} \
\end{equation}
Where $\lambda_{PDE}$, $\lambda_{BC,out}$, $\lambda_{BC,in}$, and $\lambda_{IC}$ are the weights for losses $\mathcal{L}_{PDE}$, $\mathcal{L}_{BC,out}$, $\mathcal{L}_{BC,in}$, and $\mathcal{L}_{IC}$ respectively.

\subsection{Training the PINN to approximate a solution to the heat diffusion problem}
\label{sec:method:pinn}

Since \citet{li_surrogate_2023} demonstrated that training a PINN within a subspace of learnable parameters improves accuracy and reduces training time, this paper hypothesizes that the PINN and the parameter $K$ can be optimized through a two-step process.
First, the PINN is trained to approximate the heat diffusion problem within a subspace of learnable parameters, including the estimated $K$.
Subsequently, $K$ is optimized using the trained PINN (see \cref{sec:method:klearn}), and the PINN is further refined within the same subspace of parameters, except for the updated $K$, which is adjusted accordingly.
This section focuses on how to train the PINN $\mathcal{U}$ given a fixed $K$ value.

$\tau$ and $\xi$ are randomly sampled between 0 and 1, while $K$ values are sampled from a truncated normal distribution centered around the current guess of $K$, $\hat{K}_{n}$, with a standard deviation of 0.01.
The initial value of $K$ was chosen as the midpoint of existing tabulated values for the material.
Using the previously sampled $[\tau, \xi, K]$, $\mathcal{L}_{PDE}$, $\mathcal{L}_{BC,\:out}$, and $\mathcal{L}_{BC,\:in}$ are evaluated.
$\mathcal{L}_{PDE}$ is evaluated against all $\tau$, $\xi$, and $K$ sampled, while $\mathcal{L}_{BC,\:out}$ and $\mathcal{L}_{BC,\:in}$ are evaluated against $[\tau, K]$ and $\xi$ equal to $0$ and $1$ respectively as boundary conditions apply only at the boundaries of the domain.
$\mathcal{L}_{IC}$ is evaluated at $\tau=0$ and at equally spaced intervals of $\xi$ to ensure that the entire physical domain is sampled.
To reduce the computational burden, $\mathcal{L}_{IC}(\xi)$ is evaluated on three values of $K$, which are equal to $0.95\hat{K}$, $\hat{K}$ and $1.05\hat{K}$.

The total loss is then calculated as per \cref{eq:total-loss}, the weights of $\mathcal{U}$ are updated for each step, while the weights associated with the weighted scheme of \cref{sec:method:subsec:PINN} are updated every 1000 steps.
The training is stopped when $\mathcal{L}_{PDE}$, $\mathcal{L}_{BC,out}$, $\mathcal{L}_{BC_in}$, and $\mathcal{L}_{IC}$ are lower than their respective thresholds $t_{pde}$, $t_{bc}^{in}$, $t_{bc}^{out}$, and $t_{ic}$.

\subsection{Learning $K$}
\label{sec:method:klearn}

Once the PINN is trained for a given subspace of $K$---i.e. $\mathcal{L}_{PDE}$, $\mathcal{L}_{BC,\:out}$, $\mathcal{L}_{BC,\:in}$, $\mathcal{L}_{IC}$ are all below pre-selected threshold values---$\hat{K}$ is optimized.

Given a set of $N_T$ measured thermographs $\Theta^{T}_{i}$ at times $\tau^{T}_i$ (where $\Theta^{T}_{i}$ and $\tau^{T}_i$ are the temperature and time expressed in dimensionless space), the PINN is used to predict the temperature at the wall's surface for each $\tau^{T}_i$.
Each prediction represents an estimated thermograph of the surface, which can be compared to the measured thermographs in $\Theta^{T}_{i}$.
If $\hat{K}$ is correctly estimated, the disparity between the actual and predicted thermographs should be $0$.

Thus, $\hat{K}$ can be optimized by using the loss between surface temperatures and thermographic information:
\begin{equation}
    \label{eq:TC-loss}
    \  \mathcal{L}_{TC} = \frac{1}{N_T}\sum_{i=0}^{N_{TC}} \left\vert\left\vert \mathcal{U}(W\:|\:\tau^{T}_{i},\:0,\:\hat{K}) - \Theta^{T}_{i}\right\vert\right\vert^{2}   \
\end{equation}
Based on $\mathcal{L}_{TC}$, $\hat{K}$ is updated using gradient descent, a single time for all thermographs.

Once $\hat{K}$ has been updated, the PINN is retrained within a new subspace of $K$ centered around the updated value of $\hat{K}$---as presented in \cref{sec:method:pinn}.
This process of training the PINN within a new subspace of $\hat{K}$ and optimizing $\hat{K}$ is repeated for a large enough number of steps $n$ such that $\hat{K}$ converges.

\subsection{Data collection and processing}

This section describes both the environmental data used for simulation and the generation of our ground truth measurements, as well as the implementation details of our method.
Implementation of the method and code for the evaluation is available online for reproduction.\footnote{
    \url{https://github.com/Schindler-EPFL-Lab/PINN-it}
}

\subsubsection{Weather Data}

Realistic climatic conditions are obtained from  NABEL\footnote{\url{https://www.empa.ch/web/s503/ambient-air-pollution}} (FOEN and Empa) using a weather station located in Lausanne, Switzerland, to obtain data similar to in-situ collection---i.e., close to buildings being evaluated.
Data available include solar irradiation (for $Q_{dir}$ and $Q_{diff}$ estimation), wind speed $v$ and wind direction, and ambient air temperature $T_{\inf, out}$, collected at 10-minute intervals.
In the experiments, the years 2023-2024 and 2010-2011 are used since they represented one relatively hot year in recent history and one relatively cold year in recent history.\footnote{\url{https://climate.copernicus.eu/copernicus-2023-hottest-year-record}}

\subsubsection{PINN}

The PINN implementation is done using the Jaxpi library~\cite{wang_experts_2023}.
The PINN $\mathcal{U}$ is a neural network with 4 dense layers of 256 neurons and tanh activation functions between each layer.
The optimizer used for both the PINN training and $k$ optimization is the Adam optimizer with an initial learning rate of $0.001$ and an exponential decay learning rate scheduler.

\subsubsection{Indoor and Outdoor Heat Transfer Coefficients}

The indoor heat transfer coefficient $h_{in}$ is assumed to be natural convection, which is typically within the range of 1 to 10 $W/m^{2}K$.
In the experiments, a constant value of 2 $W/m^{2}K$ was used to simplify calculations.
In principle, natural convection could be calculated in real time during PINN training.

On the other hand, the outdoor heat transfer coefficient is taken as a function of wind speed:
\begin{equation}
    \label{eq:hout-calc}
    h_{out}\;=\;18.6 \cdot v^{0.605}
\end{equation}
Indeed, \citet{evangelisti_heat_2017} showed \cref{eq:hout-calc} to be the most accurate among many other relations for an uninsulated wall.

\subsubsection{Ground Truth Generation}
\label{sec:dataset:gt}

Truth data is created using openFOAM, an open-source Finite Volume Method (FVM) software.
From the years 2023-2024 and 2010-2011, the days used for experiments consist of the first day of a month, where sufficient data was collected to be able to run an OpenFOAM simulation; a total of 25 days covering all seasons were used in the tests.

The initial condition of all openFOAM simulations was at steady state with the first environmental conditions provided.
The boundary conditions were supplied by weather data obtained from NABEL weather stations.
The data was transformed into sol-air temperature and outdoor heat transfer coefficient using equations \cref{eq:solair} and \cref{eq:hout-calc}, assuming 100\% of the sunlight was diffuse and that surface albedo is 1.
The indoor ambient air temperature and heat transfer coefficient were kept constant during simulations using values of 298.15 K and 2 W/m$^{2}$, respectively.
In all experiments, ground-truth thermographic data consists of temperature readings on the outer surface of the wall over the time of the simulation.

\subsubsection{Hardware}

All experiments were run on an Nvidia T4 GPU.

\subsubsection{Statistics and reproducibility}

Given the limited number of samples (one for each day, so 24), confidence intervals were obtained using bootstrapping by taking 1000 samples 10,000 times.
To reproduce the primary results of this research, refer to the analytical pipeline available at \url{https://github.com/Schindler-EPFL-Lab/PINN-it}.

\section{Data availability}

The unprocessed data are available with the code at \url{https://github.com/Schindler-EPFL-Lab/PINN-it}.
The environmental data can be obtained by contacting NABEL.

\section{Code availability}

The source code for PINN-it is available at: \url{https://github.com/Schindler-EPFL-Lab/PINN-it}.

\bibliography{AliProject,references,PINN-it}%


\begin{thebibliography}{46}
\ifx \bisbn   \undefined \def \bisbn  #1{ISBN #1}\fi
\ifx \binits  \undefined \def \binits#1{#1}\fi
\ifx \bauthor  \undefined \def \bauthor#1{#1}\fi
\ifx \batitle  \undefined \def \batitle#1{#1}\fi
\ifx \bjtitle  \undefined \def \bjtitle#1{#1}\fi
\ifx \bvolume  \undefined \def \bvolume#1{\textbf{#1}}\fi
\ifx \byear  \undefined \def \byear#1{#1}\fi
\ifx \bissue  \undefined \def \bissue#1{#1}\fi
\ifx \bfpage  \undefined \def \bfpage#1{#1}\fi
\ifx \blpage  \undefined \def \blpage #1{#1}\fi
\ifx \burl  \undefined \def \burl#1{\textsf{#1}}\fi
\ifx \doiurl  \undefined \def \doiurl#1{\url{https://doi.org/#1}}\fi
\ifx \betal  \undefined \def \betal{\textit{et al.}}\fi
\ifx \binstitute  \undefined \def \binstitute#1{#1}\fi
\ifx \binstitutionaled  \undefined \def \binstitutionaled#1{#1}\fi
\ifx \bctitle  \undefined \def \bctitle#1{#1}\fi
\ifx \beditor  \undefined \def \beditor#1{#1}\fi
\ifx \bpublisher  \undefined \def \bpublisher#1{#1}\fi
\ifx \bbtitle  \undefined \def \bbtitle#1{#1}\fi
\ifx \bedition  \undefined \def \bedition#1{#1}\fi
\ifx \bseriesno  \undefined \def \bseriesno#1{#1}\fi
\ifx \blocation  \undefined \def \blocation#1{#1}\fi
\ifx \bsertitle  \undefined \def \bsertitle#1{#1}\fi
\ifx \bsnm \undefined \def \bsnm#1{#1}\fi
\ifx \bsuffix \undefined \def \bsuffix#1{#1}\fi
\ifx \bparticle \undefined \def \bparticle#1{#1}\fi
\ifx \barticle \undefined \def \barticle#1{#1}\fi
\bibcommenthead
\ifx \bconfdate \undefined \def \bconfdate #1{#1}\fi
\ifx \botherref \undefined \def \botherref #1{#1}\fi
\ifx \url \undefined \def \url#1{\textsf{#1}}\fi
\ifx \bchapter \undefined \def \bchapter#1{#1}\fi
\ifx \bbook \undefined \def \bbook#1{#1}\fi
\ifx \bcomment \undefined \def \bcomment#1{#1}\fi
\ifx \oauthor \undefined \def \oauthor#1{#1}\fi
\ifx \citeauthoryear \undefined \def \citeauthoryear#1{#1}\fi
\ifx \endbibitem  \undefined \def \endbibitem {}\fi
\ifx \bconflocation  \undefined \def \bconflocation#1{#1}\fi
\ifx \arxivurl  \undefined \def \arxivurl#1{\textsf{#1}}\fi
\csname PreBibitemsHook\endcsname

\bibitem[\protect\citeauthoryear{Wang and Wang}{2004}]{wang2004pde}
\begin{barticle}
\bauthor{\bsnm{Wang}, \binits{M.Y.}},
\bauthor{\bsnm{Wang}, \binits{X.}}:
\batitle{Pde-driven level sets, shape sensitivity and curvature flow for structural topology optimization}.
\bjtitle{Computer Modeling in Engineering \& Sciences}
\bvolume{6}(\bissue{4}),
\bfpage{373}
(\byear{2004})
\end{barticle}
\endbibitem

\bibitem[\protect\citeauthoryear{Wallin et~al.}{2020}]{wallin2020Consis}
\begin{barticle}
\bauthor{\bsnm{Wallin}, \binits{M.}},
\bauthor{\bsnm{Ivarsson}, \binits{N.}},
\bauthor{\bsnm{Amir}, \binits{O.}},
\bauthor{\bsnm{Tortorelli}, \binits{D.}}:
\batitle{Consistent boundary conditions for pde filter regularization in topology optimization}.
\bjtitle{Structural and Multidisciplinary Optimization}
\bvolume{62}(\bissue{3}),
\bfpage{1299}--\blpage{1311}
(\byear{2020})
\doiurl{10.1007/s00158-020-02556-w}
\end{barticle}
\endbibitem

\bibitem[\protect\citeauthoryear{Ugail et~al.}{1999}]{ugail1999Techni}
\begin{barticle}
\bauthor{\bsnm{Ugail}, \binits{H.}},
\bauthor{\bsnm{Bloor}, \binits{M.I.G.}},
\bauthor{\bsnm{Wilson}, \binits{M.J.}}:
\batitle{Techniques for interactive design using the pde method}.
\bjtitle{ACM Transactions on Graphics}
\bvolume{18}(\bissue{2}),
\bfpage{195}--\blpage{212}
(\byear{1999})
\doiurl{10.1145/318009.318078}
\end{barticle}
\endbibitem

\bibitem[\protect\citeauthoryear{Ferziger et~al.}{2020}]{ferziger2020Comput}
\begin{botherref}
\oauthor{\bsnm{Ferziger}, \binits{J.H.}},
\oauthor{\bsnm{Perić}, \binits{M.}},
\oauthor{\bsnm{Street}, \binits{R.L.}}:
Computational methods for fluid dynamics
(2020)
\doiurl{10.1007/978-3-319-99693-6}
\end{botherref}
\endbibitem

\bibitem[\protect\citeauthoryear{Ansari et~al.}{2022}]{ansari2022Solvin}
\begin{barticle}
\bauthor{\bsnm{Ansari}, \binits{A.}},
\bauthor{\bsnm{Pasyar}, \binits{A.}},
\bauthor{\bsnm{Ghorbani}, \binits{A.}}:
\batitle{Solving 2-d gravity inversion problems using a pde model in geophysics exploration}.
\bjtitle{International Journal of Mining and Geo-Engineering}
(\byear{2022})
\doiurl{10.22059/ijmge.2021.317502.594888}
\end{barticle}
\endbibitem

\bibitem[\protect\citeauthoryear{Bagtzoglou and Baun~†}{2005}]{bagtzoglou2005Near_r}
\begin{barticle}
\bauthor{\bsnm{Bagtzoglou}, \binits{A.C.}},
\bauthor{\bsnm{Baun~†}, \binits{S.A.}}:
\batitle{Near real-time atmospheric contamination source identification by an optimization-based inverse method}.
\bjtitle{Inverse Problems in Science and Engineering}
\bvolume{13}(\bissue{3}),
\bfpage{241}--\blpage{259}
(\byear{2005})
\doiurl{10.1080/10682760412331330163}
\end{barticle}
\endbibitem

\bibitem[\protect\citeauthoryear{Nolte and Bertoglio}{2022}]{nolte2022Invers}
\begin{botherref}
\oauthor{\bsnm{Nolte}, \binits{D.}},
\oauthor{\bsnm{Bertoglio}, \binits{C.}}:
Inverse problems in blood flow modeling: A review.
International Journal for Numerical Methods in Biomedical Engineering
\textbf{38}(8)
(2022)
\doiurl{10.1002/cnm.3613}
\end{botherref}
\endbibitem

\bibitem[\protect\citeauthoryear{Fischer et~al.}{2020}]{fischer2020Scalab}
\begin{barticle}
\bauthor{\bsnm{Fischer}, \binits{P.}},
\bauthor{\bsnm{Min}, \binits{M.}},
\bauthor{\bsnm{Rathnayake}, \binits{T.}},
\bauthor{\bsnm{Dutta}, \binits{S.}},
\bauthor{\bsnm{Kolev}, \binits{T.}},
\bauthor{\bsnm{Dobrev}, \binits{V.}},
\bauthor{\bsnm{Camier}, \binits{J.-S.}},
\bauthor{\bsnm{Kronbichler}, \binits{M.}},
\bauthor{\bsnm{Warburton}, \binits{T.}},
\bauthor{\bsnm{Świrydowicz}, \binits{K.}},
\bauthor{\bsnm{Brown}, \binits{J.}}:
\batitle{Scalability of high-performance pde solvers}.
\bjtitle{The International Journal of High Performance Computing Applications}
\bvolume{34}(\bissue{5}),
\bfpage{562}--\blpage{586}
(\byear{2020})
\doiurl{10.1177/1094342020915762}
\end{barticle}
\endbibitem

\bibitem[\protect\citeauthoryear{Cardiff and Kitanidis}{2008}]{cardiff2008Effici}
\begin{barticle}
\bauthor{\bsnm{Cardiff}, \binits{M.}},
\bauthor{\bsnm{Kitanidis}, \binits{P.K.}}:
\batitle{Efficient solution of nonlinear, underdetermined inverse problems with a generalized pde model}.
\bjtitle{Computers and Geosciences}
\bvolume{34}(\bissue{11}),
\bfpage{1480}--\blpage{1491}
(\byear{2008})
\doiurl{10.1016/j.cageo.2008.01.013}
\end{barticle}
\endbibitem

\bibitem[\protect\citeauthoryear{Schmidtke et~al.}{2023}]{10178253}
\begin{bchapter}
\bauthor{\bsnm{Schmidtke}, \binits{V.}},
\bauthor{\bsnm{Rüger}, \binits{M.}},
\bauthor{\bsnm{Stursberg}, \binits{O.}}:
\bctitle{Model predictive control of pdes for temperature control in 3d-printing processes}.
In: \bbtitle{2023 European Control Conference (ECC)},
pp. \bfpage{1}--\blpage{6}
(\byear{2023}).
\doiurl{10.23919/ECC57647.2023.10178253}
\end{bchapter}
\endbibitem

\bibitem[\protect\citeauthoryear{Zheng et~al.}{2017}]{8264402}
\begin{bchapter}
\bauthor{\bsnm{Zheng}, \binits{C.}},
\bauthor{\bsnm{Wen}, \binits{J.T.}},
\bauthor{\bsnm{Mishra}, \binits{S.}},
\bauthor{\bsnm{Diagne}, \binits{M.}}:
\bctitle{Modeling and cooling rate control in laser additive manufacturing: 1-d pde formulation}.
In: \bbtitle{2017 IEEE 56th Annual Conference on Decision and Control (CDC)},
pp. \bfpage{5020}--\blpage{5025}
(\byear{2017}).
\doiurl{10.1109/CDC.2017.8264402}
\end{bchapter}
\endbibitem

\bibitem[\protect\citeauthoryear{Dong et~al.}{2018}]{8310645}
\begin{barticle}
\bauthor{\bsnm{Dong}, \binits{J.}},
\bauthor{\bsnm{Wang}, \binits{Q.}},
\bauthor{\bsnm{Wang}, \binits{M.}},
\bauthor{\bsnm{Peng}, \binits{K.}}:
\batitle{Data-driven quality monitoring techniques for distributed parameter systems with application to hot-rolled strip laminar cooling process}.
\bjtitle{IEEE Access}
\bvolume{6},
\bfpage{16646}--\blpage{16654}
(\byear{2018})
\doiurl{10.1109/ACCESS.2018.2812919}
\end{barticle}
\endbibitem

\bibitem[\protect\citeauthoryear{Nardi et~al.}{2018}]{Nardi_2018}
\begin{barticle}
\bauthor{\bsnm{Nardi}, \binits{I.}},
\bauthor{\bsnm{Lucchi}, \binits{E.}},
\bauthor{\bsnm{Rubeis}, \binits{T.}},
\bauthor{\bsnm{Ambrosini}, \binits{D.}}:
\batitle{Quantification of heat energy losses through the building envelope: A state-of-the-art analysis with critical and comprehensive review on infrared thermography}.
\bjtitle{Building and Environment}
\bvolume{146},
\bfpage{190}--\blpage{205}
(\byear{2018})
\doiurl{10.1016/j.buildenv.2018.09.050}
\end{barticle}
\endbibitem

\bibitem[\protect\citeauthoryear{Sarihi et~al.}{2021}]{SARIHI2021102525}
\begin{barticle}
\bauthor{\bsnm{Sarihi}, \binits{S.}},
\bauthor{\bsnm{{Mehdizadeh Saradj}}, \binits{F.}},
\bauthor{\bsnm{Faizi}, \binits{M.}}:
\batitle{A critical review of façade retrofit measures for minimizing heating and cooling demand in existing buildings}.
\bjtitle{Sustainable Cities and Society}
\bvolume{64},
\bfpage{102525}
(\byear{2021})
\doiurl{10.1016/j.scs.2020.102525}
\end{barticle}
\endbibitem

\bibitem[\protect\citeauthoryear{Commission}{2020}]{act2011communication}
\begin{botherref}
\oauthor{\bsnm{Commission}, \binits{E.}}:
A renovation wave for europe—greening our buildings, creating jobs, improving lives.
Official Journal of the European Union,
26
(2020)
\end{botherref}
\endbibitem

\bibitem[\protect\citeauthoryear{Tejedor et~al.}{2017}]{tejedor_quantitative_2017}
\begin{barticle}
\bauthor{\bsnm{Tejedor}, \binits{B.}},
\bauthor{\bsnm{Casals}, \binits{M.}},
\bauthor{\bsnm{Gangolells}, \binits{M.}},
\bauthor{\bsnm{Roca}, \binits{X.}}:
\batitle{Quantitative internal infrared thermography for determining in-situ thermal behaviour of façades}.
\bjtitle{Energy and Buildings}
\bvolume{151},
\bfpage{187}--\blpage{197}
(\byear{2017})
\doiurl{10.1016/j.enbuild.2017.06.040} .
Accessed 2024-04-16
\end{barticle}
\endbibitem

\bibitem[\protect\citeauthoryear{}{}]{noauthor_iso_nodate}
\begin{botherref}
{ISO} 6946:2017.
\url{https://www.iso.org/standard/65708.html}
Accessed 2025-05-13
\end{botherref}
\endbibitem

\bibitem[\protect\citeauthoryear{}{}]{noauthor_iso_nodate-1}
\begin{botherref}
{ISO} 9869-1:2014.
\url{https://www.iso.org/standard/59697.html}
Accessed 2025-05-14
\end{botherref}
\endbibitem

\bibitem[\protect\citeauthoryear{Tardy}{2023}]{tardy_review_2023}
\begin{barticle}
\bauthor{\bsnm{Tardy}, \binits{F.}}:
\batitle{A review of the use of infrared thermography in building envelope thermal property characterization studies}.
\bjtitle{Journal of Building Engineering}
\bvolume{75},
\bfpage{106918}
(\byear{2023})
\doiurl{10.1016/j.jobe.2023.106918} .
Accessed 2025-05-14
\end{barticle}
\endbibitem

\bibitem[\protect\citeauthoryear{Desogus et~al.}{2011}]{desogus_comparing_2011}
\begin{barticle}
\bauthor{\bsnm{Desogus}, \binits{G.}},
\bauthor{\bsnm{Mura}, \binits{S.}},
\bauthor{\bsnm{Ricciu}, \binits{R.}}:
\batitle{Comparing different approaches to in situ measurement of building components thermal resistance}.
\bjtitle{Energy and Buildings}
\bvolume{43}(\bissue{10}),
\bfpage{2613}--\blpage{2620}
(\byear{2011})
\doiurl{10.1016/j.enbuild.2011.05.025} .
Accessed 2025-05-14
\end{barticle}
\endbibitem

\bibitem[\protect\citeauthoryear{Cesaratto et~al.}{2011}]{cesaratto_effect_2011}
\begin{barticle}
\bauthor{\bsnm{Cesaratto}, \binits{P.G.}},
\bauthor{\bsnm{De~Carli}, \binits{M.}},
\bauthor{\bsnm{Marinetti}, \binits{S.}}:
\batitle{Effect of different parameters on the in situ thermal conductance evaluation}.
\bjtitle{Energy and Buildings}
\bvolume{43}(\bissue{7}),
\bfpage{1792}--\blpage{1801}
(\byear{2011})
\doiurl{10.1016/j.enbuild.2011.03.021} .
Accessed 2025-05-14
\end{barticle}
\endbibitem

\bibitem[\protect\citeauthoryear{Trethowen}{1986}]{trethowen_measurement_1986}
\begin{barticle}
\bauthor{\bsnm{Trethowen}, \binits{H.}}:
\batitle{Measurement errors with surface-mounted heat flux sensors}.
\bjtitle{Building and Environment}
\bvolume{21}(\bissue{1}),
\bfpage{41}--\blpage{56}
(\byear{1986})
\doiurl{10.1016/0360-1323(86)90007-7} .
Accessed 2025-05-14
\end{barticle}
\endbibitem

\bibitem[\protect\citeauthoryear{Peng and Wu}{2008}]{peng_situ_2008}
\begin{barticle}
\bauthor{\bsnm{Peng}, \binits{C.}},
\bauthor{\bsnm{Wu}, \binits{Z.}}:
\batitle{In situ measuring and evaluating the thermal resistance of building construction}.
\bjtitle{Energy and Buildings}
\bvolume{40}(\bissue{11}),
\bfpage{2076}--\blpage{2082}
(\byear{2008})
\doiurl{10.1016/j.enbuild.2008.05.012} .
Accessed 2025-05-14
\end{barticle}
\endbibitem

\bibitem[\protect\citeauthoryear{Evangelisti et~al.}{2020}]{evangelisti_methodological_2020}
\begin{barticle}
\bauthor{\bsnm{Evangelisti}, \binits{L.}},
\bauthor{\bsnm{Guattari}, \binits{C.}},
\bauthor{\bsnm{De~Lieto~Vollaro}, \binits{R.}},
\bauthor{\bsnm{Asdrubali}, \binits{F.}}:
\batitle{A methodological approach for heat-flow meter data post-processing under different climatic conditions and wall orientations}.
\bjtitle{Energy and Buildings}
\bvolume{223},
\bfpage{110216}
(\byear{2020})
\doiurl{10.1016/j.enbuild.2020.110216} .
Accessed 2025-05-13
\end{barticle}
\endbibitem

\bibitem[\protect\citeauthoryear{Biddulph et~al.}{2014}]{biddulph_inferring_2014}
\begin{barticle}
\bauthor{\bsnm{Biddulph}, \binits{P.}},
\bauthor{\bsnm{Gori}, \binits{V.}},
\bauthor{\bsnm{Elwell}, \binits{C.A.}},
\bauthor{\bsnm{Scott}, \binits{C.}},
\bauthor{\bsnm{Rye}, \binits{C.}},
\bauthor{\bsnm{Lowe}, \binits{R.}},
\bauthor{\bsnm{Oreszczyn}, \binits{T.}}:
\batitle{Inferring the thermal resistance and effective thermal mass of a wall using frequent temperature and heat flux measurements}.
\bjtitle{Energy and Buildings}
\bvolume{78},
\bfpage{10}--\blpage{16}
(\byear{2014})
\doiurl{10.1016/j.enbuild.2014.04.004} .
Accessed 2025-05-14
\end{barticle}
\endbibitem

\bibitem[\protect\citeauthoryear{Tejedor et~al.}{2018}]{tejedor_assessing_2018}
\begin{barticle}
\bauthor{\bsnm{Tejedor}, \binits{B.}},
\bauthor{\bsnm{Casals}, \binits{M.}},
\bauthor{\bsnm{Gangolells}, \binits{M.}}:
\batitle{Assessing the influence of operating conditions and thermophysical properties on the accuracy of in-situ measured {U} -values using quantitative internal infrared thermography}.
\bjtitle{Energy and Buildings}
\bvolume{171},
\bfpage{64}--\blpage{75}
(\byear{2018})
\doiurl{10.1016/j.enbuild.2018.04.011} .
Accessed 2025-05-14
\end{barticle}
\endbibitem

\bibitem[\protect\citeauthoryear{Lu and Memari}{}]{lu_application_2019}
\begin{botherref}
\oauthor{\bsnm{Lu}, \binits{X.}},
\oauthor{\bsnm{Memari}, \binits{A.}}:
Application of infrared thermography for in-situ determination of building envelope thermal properties
\textbf{26},
100885
\doiurl{10.1016/j.jobe.2019.100885} .
Accessed 2024-04-16
\end{botherref}
\endbibitem

\bibitem[\protect\citeauthoryear{Mahmoodzadeh et~al.}{2021}]{mahmoodzadeh_determining_2021}
\begin{barticle}
\bauthor{\bsnm{Mahmoodzadeh}, \binits{M.}},
\bauthor{\bsnm{Gretka}, \binits{V.}},
\bauthor{\bsnm{Hay}, \binits{K.}},
\bauthor{\bsnm{Steele}, \binits{C.}},
\bauthor{\bsnm{Mukhopadhyaya}, \binits{P.}}:
\batitle{Determining overall heat transfer coefficient ({U}-{Value}) of wood-framed wall assemblies in {Canada} using external infrared thermography}.
\bjtitle{Building and Environment}
\bvolume{199},
\bfpage{107897}
(\byear{2021})
\doiurl{10.1016/j.buildenv.2021.107897} .
Accessed 2025-05-14
\end{barticle}
\endbibitem

\bibitem[\protect\citeauthoryear{Marino et~al.}{2017}]{marino_estimation_2017}
\begin{barticle}
\bauthor{\bsnm{Marino}, \binits{B.M.}},
\bauthor{\bsnm{Muñoz}, \binits{N.}},
\bauthor{\bsnm{Thomas}, \binits{L.P.}}:
\batitle{Estimation of the surface thermal resistances and heat loss by conduction using thermography}.
\bjtitle{Applied Thermal Engineering}
\bvolume{114},
\bfpage{1213}--\blpage{1221}
(\byear{2017})
\doiurl{10.1016/j.applthermaleng.2016.12.033} .
Accessed 2025-05-14
\end{barticle}
\endbibitem

\bibitem[\protect\citeauthoryear{González-Aguilera et~al.}{2013}]{GONZALEZAGUILERA201329}
\begin{barticle}
\bauthor{\bsnm{González-Aguilera}, \binits{D.}},
\bauthor{\bsnm{Lagüela}, \binits{S.}},
\bauthor{\bsnm{Rodríguez-Gonzálvez}, \binits{P.}},
\bauthor{\bsnm{Hernández-López}, \binits{D.}}:
\batitle{Image-based thermographic modeling for assessing energy efficiency of buildings façades}.
\bjtitle{Energy and Buildings}
\bvolume{65},
\bfpage{29}--\blpage{36}
(\byear{2013})
\doiurl{10.1016/j.enbuild.2013.05.040}
\end{barticle}
\endbibitem

\bibitem[\protect\citeauthoryear{{Videras Rodríguez} et~al.}{2024}]{VIDERASRODRIGUEZ2024114874}
\begin{barticle}
\bauthor{\bsnm{{Videras Rodríguez}}, \binits{M.}},
\bauthor{\bsnm{{Gómez Melgar}}, \binits{S.}},
\bauthor{\bsnm{{Andújar Márquez}}, \binits{J.M.}}:
\batitle{Evaluation of aerial thermography for measuring the thermal transmittance (u-value) of a building façade}.
\bjtitle{Energy and Buildings}
\bvolume{324},
\bfpage{114874}
(\byear{2024})
\doiurl{10.1016/j.enbuild.2024.114874}
\end{barticle}
\endbibitem

\bibitem[\protect\citeauthoryear{Albatici et~al.}{2015}]{ALBATICI2015218}
\begin{barticle}
\bauthor{\bsnm{Albatici}, \binits{R.}},
\bauthor{\bsnm{Tonelli}, \binits{A.M.}},
\bauthor{\bsnm{Chiogna}, \binits{M.}}:
\batitle{A comprehensive experimental approach for the validation of quantitative infrared thermography in the evaluation of building thermal transmittance}.
\bjtitle{Applied Energy}
\bvolume{141},
\bfpage{218}--\blpage{228}
(\byear{2015})
\doiurl{10.1016/j.apenergy.2014.12.035}
\end{barticle}
\endbibitem

\bibitem[\protect\citeauthoryear{Bacher and Madsen}{2011}]{bacher_identifying_2011}
\begin{barticle}
\bauthor{\bsnm{Bacher}, \binits{P.}},
\bauthor{\bsnm{Madsen}, \binits{H.}}:
\batitle{Identifying suitable models for the heat dynamics of buildings}.
\bjtitle{Energy and Buildings}
\bvolume{43}(\bissue{7}),
\bfpage{1511}--\blpage{1522}
(\byear{2011})
\doiurl{10.1016/j.enbuild.2011.02.005} .
Accessed 2025-05-14
\end{barticle}
\endbibitem

\bibitem[\protect\citeauthoryear{Chassaing et~al.}{2025}]{chassaing2025thermoxels}
\begin{botherref}
\oauthor{\bsnm{Chassaing}, \binits{E.}},
\oauthor{\bsnm{Forest}, \binits{F.}},
\oauthor{\bsnm{Fink}, \binits{O.}},
\oauthor{\bsnm{Mielle}, \binits{M.}}:
Thermoxels: a voxel-based method to generate simulation-ready 3d thermal models.
arXiv preprint arXiv:2504.04448
(2025)
\end{botherref}
\endbibitem

\bibitem[\protect\citeauthoryear{Raissi et~al.}{2019}]{raissi_physics-informed_2019}
\begin{barticle}
\bauthor{\bsnm{Raissi}, \binits{M.}},
\bauthor{\bsnm{Perdikaris}, \binits{P.}},
\bauthor{\bsnm{Karniadakis}, \binits{G.E.}}:
\batitle{Physics-informed neural networks: {A} deep learning framework for solving forward and inverse problems involving nonlinear partial differential equations}.
\bjtitle{Journal of Computational physics}
\bvolume{378},
\bfpage{686}--\blpage{707}
(\byear{2019}).
\bcomment{Publisher: Elsevier}.
Accessed 2025-05-14
\end{barticle}
\endbibitem

\bibitem[\protect\citeauthoryear{Lawal et~al.}{2022}]{bdcc6040140}
\begin{botherref}
\oauthor{\bsnm{Lawal}, \binits{Z.K.}},
\oauthor{\bsnm{Yassin}, \binits{H.}},
\oauthor{\bsnm{Lai}, \binits{D.T.C.}},
\oauthor{\bsnm{Che~Idris}, \binits{A.}}:
Physics-informed neural network (pinn) evolution and beyond: A systematic literature review and bibliometric analysis.
Big Data and Cognitive Computing
\textbf{6}(4)
(2022)
\doiurl{10.3390/bdcc6040140}
\end{botherref}
\endbibitem

\bibitem[\protect\citeauthoryear{Hu and Kabala}{2023}]{hu2023predicting}
\begin{barticle}
\bauthor{\bsnm{Hu}, \binits{A.V.}},
\bauthor{\bsnm{Kabala}, \binits{Z.J.}}:
\batitle{Predicting and reconstructing aerosol--cloud--precipitation interactions with physics-informed neural networks}.
\bjtitle{Atmosphere}
\bvolume{14}(\bissue{12}),
\bfpage{1798}
(\byear{2023})
\end{barticle}
\endbibitem

\bibitem[\protect\citeauthoryear{Millevoi et~al.}{2024}]{millevoi2024physics}
\begin{barticle}
\bauthor{\bsnm{Millevoi}, \binits{C.}},
\bauthor{\bsnm{Pasetto}, \binits{D.}},
\bauthor{\bsnm{Ferronato}, \binits{M.}}:
\batitle{A physics-informed neural network approach for compartmental epidemiological models}.
\bjtitle{PLOS Computational Biology}
\bvolume{20}(\bissue{9}),
\bfpage{1012387}
(\byear{2024})
\end{barticle}
\endbibitem

\bibitem[\protect\citeauthoryear{Chiu et~al.}{2022}]{chiu2022can}
\begin{barticle}
\bauthor{\bsnm{Chiu}, \binits{P.-H.}},
\bauthor{\bsnm{Wong}, \binits{J.C.}},
\bauthor{\bsnm{Ooi}, \binits{C.}},
\bauthor{\bsnm{Dao}, \binits{M.H.}},
\bauthor{\bsnm{Ong}, \binits{Y.-S.}}:
\batitle{Can-pinn: A fast physics-informed neural network based on coupled-automatic--numerical differentiation method}.
\bjtitle{Computer Methods in Applied Mechanics and Engineering}
\bvolume{395},
\bfpage{114909}
(\byear{2022})
\end{barticle}
\endbibitem

\bibitem[\protect\citeauthoryear{Labib}{2025}]{labib_utilizing_2025}
\begin{barticle}
\bauthor{\bsnm{Labib}, \binits{R.}}:
\batitle{Utilizing physics-informed neural networks to advance daylighting simulations in buildings}.
\bjtitle{Journal of Building Engineering}
\bvolume{100},
\bfpage{111726}
(\byear{2025})
\doiurl{10.1016/j.jobe.2024.111726} .
Accessed 2025-05-13
\end{barticle}
\endbibitem

\bibitem[\protect\citeauthoryear{Cai et~al.}{2021}]{cai_physics-informed_2021}
\begin{barticle}
\bauthor{\bsnm{Cai}, \binits{S.}},
\bauthor{\bsnm{Wang}, \binits{Z.}},
\bauthor{\bsnm{Wang}, \binits{S.}},
\bauthor{\bsnm{Perdikaris}, \binits{P.}},
\bauthor{\bsnm{Karniadakis}, \binits{G.E.}}:
\batitle{Physics-{Informed} {Neural} {Networks} for {Heat} {Transfer} {Problems}}.
\bjtitle{Journal of Heat Transfer}
\bvolume{143}(\bissue{6}),
\bfpage{060801}
(\byear{2021})
\doiurl{10.1115/1.4050542} .
Accessed 2025-05-13
\end{barticle}
\endbibitem

\bibitem[\protect\citeauthoryear{Zobeiry and Humfeld}{2021}]{zobeiry_physics-informed_2021}
\begin{barticle}
\bauthor{\bsnm{Zobeiry}, \binits{N.}},
\bauthor{\bsnm{Humfeld}, \binits{K.D.}}:
\batitle{A physics-informed machine learning approach for solving heat transfer equation in advanced manufacturing and engineering applications}.
\bjtitle{Engineering Applications of Artificial Intelligence}
\bvolume{101},
\bfpage{104232}
(\byear{2021})
\doiurl{10.1016/j.engappai.2021.104232} .
Accessed 2024-04-18
\end{barticle}
\endbibitem

\bibitem[\protect\citeauthoryear{Billah et~al.}{2023}]{billah_physics-informed_2023}
\begin{barticle}
\bauthor{\bsnm{Billah}, \binits{M.M.}},
\bauthor{\bsnm{Khan}, \binits{A.I.}},
\bauthor{\bsnm{Liu}, \binits{J.}},
\bauthor{\bsnm{Dutta}, \binits{P.}}:
\batitle{Physics-informed deep neural network for inverse heat transfer problems in materials}.
\bjtitle{Materials Today Communications}
\bvolume{35},
\bfpage{106336}
(\byear{2023})
\doiurl{10.1016/j.mtcomm.2023.106336} .
Accessed 2025-05-14
\end{barticle}
\endbibitem

\bibitem[\protect\citeauthoryear{Wang et~al.}{2023}]{wang_experts_2023}
\begin{botherref}
\oauthor{\bsnm{Wang}, \binits{S.}},
\oauthor{\bsnm{Sankaran}, \binits{S.}},
\oauthor{\bsnm{Wang}, \binits{H.}},
\oauthor{\bsnm{Perdikaris}, \binits{P.}}:
An {Expert}'s {Guide} to {Training} {Physics}-informed {Neural} {Networks}.
arXiv.
arXiv:2308.08468 [cs]
(2023).
\doiurl{10.48550/arXiv.2308.08468} .
\url{http://arxiv.org/abs/2308.08468}
Accessed 2025-05-14
\end{botherref}
\endbibitem

\bibitem[\protect\citeauthoryear{Li et~al.}{2023}]{li_surrogate_2023}
\begin{barticle}
\bauthor{\bsnm{Li}, \binits{Y.}},
\bauthor{\bsnm{Wang}, \binits{Y.}},
\bauthor{\bsnm{Yan}, \binits{L.}}:
\batitle{Surrogate modeling for {Bayesian} inverse problems based on physics-informed neural networks}.
\bjtitle{Journal of Computational Physics}
\bvolume{475},
\bfpage{111841}
(\byear{2023})
\doiurl{10.1016/j.jcp.2022.111841} .
Accessed 2025-05-14
\end{barticle}
\endbibitem

\bibitem[\protect\citeauthoryear{Evangelisti et~al.}{2017}]{evangelisti_heat_2017}
\begin{barticle}
\bauthor{\bsnm{Evangelisti}, \binits{L.}},
\bauthor{\bsnm{Guattari}, \binits{C.}},
\bauthor{\bsnm{Gori}, \binits{P.}},
\bauthor{\bsnm{Bianchi}, \binits{F.}}:
\batitle{Heat transfer study of external convective and radiative coefficients for building applications}.
\bjtitle{Energy and Buildings}
\bvolume{151},
\bfpage{429}--\blpage{438}
(\byear{2017})
\doiurl{10.1016/j.enbuild.2017.07.004} .
Accessed 2025-05-14
\end{barticle}
\endbibitem

\end{thebibliography}

\end{document}